\documentclass[letterpaper]{article} 
\usepackage{aaai2026}
\usepackage{times}  
\usepackage{helvet}  
\usepackage{courier}  
\usepackage[hyphens]{url}  
\usepackage{graphicx} 
\usepackage{amsmath}
\usepackage{amssymb}
\usepackage{dsfont}
\usepackage{xcolor}
\usepackage{booktabs}
\usepackage{multirow}
\usepackage{multicol}
\usepackage{colortbl, booktabs}
\usepackage{natbib}  
\usepackage{caption} 
\usepackage{algorithm}
\usepackage{algorithmic}

\usepackage{tcolorbox}
\usepackage{makecell}
\usepackage{bbding}

\usepackage{newfloat}
\usepackage{listings}
\DeclareCaptionStyle{ruled}{labelfont=normalfont,labelsep=colon,strut=off} 
\floatstyle{ruled}
\newfloat{listing}{tb}{lst}{}
\floatname{listing}{Listing}
\nocopyright

\title{MedMKEB: A Comprehensive Knowledge Editing Benchmark for Medical Multimodal Large Language Models}
\author{
    Dexuan Xu\textsuperscript{\rm 1},
    Jieyi Wang\textsuperscript{\rm 2},
    Zhongyan Chai\textsuperscript{\rm 2},
    Yongzhi Cao\textsuperscript{\rm 1},
    Hanpin Wang\textsuperscript{\rm 1},\\
    Huamin Zhang\textsuperscript{\rm 3},
    Yu Huang\textsuperscript{\rm 4}\thanks{Corresponding Author}
}
\affiliations{
    \textsuperscript{\rm 1}School of Computer Science, Peking University\\
    \textsuperscript{\rm 2}School of Software and Microelectronics, Peking University\\
    \textsuperscript{\rm 3}Institute of Basic Theory of Chinese Medicine, China Academy of Chinese Medical Sciences\\
    \textsuperscript{\rm 4}National Engineering Research Center For Software Engineering, Peking University

}

\usepackage{bibentry}

\begin{document}

\maketitle

\begin{abstract}
Recent advances in multimodal large language models (MLLMs) have significantly improved medical AI, enabling it to unify the understanding of visual and textual information. However, as medical knowledge continues to evolve, it is critical to allow these models to efficiently update outdated or incorrect information without retraining from scratch. Although textual knowledge editing has been widely studied, there is still a lack of systematic benchmarks for multimodal medical knowledge editing involving image and text modalities. To fill this gap, we present MedMKEB, the first comprehensive benchmark designed to evaluate the reliability, generality, locality, portability, and robustness of knowledge editing in medical multimodal large language models. MedMKEB is built on a high-quality medical visual question-answering dataset and enriched with carefully constructed editing tasks, including counterfactual correction, semantic generalization, knowledge transfer, and adversarial robustness. We incorporate human expert validation to ensure the accuracy and reliability of the benchmark. Extensive single editing and sequential editing experiments on state-of-the-art general and medical MLLMs demonstrate the limitations of existing knowledge-based editing approaches in medicine, highlighting the need to develop specialized editing strategies. MedMKEB will serve as a standard benchmark to promote the development of trustworthy and efficient medical knowledge editing algorithms.
\end{abstract}
\section{Introduction}

Medical Multimodal Large Language Models~(Medical MLLMs) have become powerful tools with the ability to answer clinical questions, interpret medical images, and support a variety of medical decisions~\cite{xiao2025comprehensive}. Such models are usually trained on large-scale medical image-text pairs and can capture the complex relationship between vision and language. However, when medical facts change, how to accurately and locally edit existing knowledge in the model without affecting its overall performance remains a key issue that has not been fully explored~\cite{xu2025knowledge}.

Knowledge editing is a method that can update, modify, or delete model-specific knowledge without retraining the model. It has become a research hotspot in natural language processing in recent years. Existing work mainly focuses on plain text language models and benchmark datasets, such as ZsRE~\cite{levy2017zero} and CounterFact~\cite{meng2022locating}. However, in the medical field, knowledge is inherently multimodal: medical judgments often rely on the comprehensive analysis of visual evidence, such as radiological images and pathological images, as well as professional text information. Therefore, the multimodal medical knowledge editing task urgently needs new task definitions, benchmark designs, and evaluation methods.

\begin{figure}[!t]
\centerline{\includegraphics[width=1\linewidth]{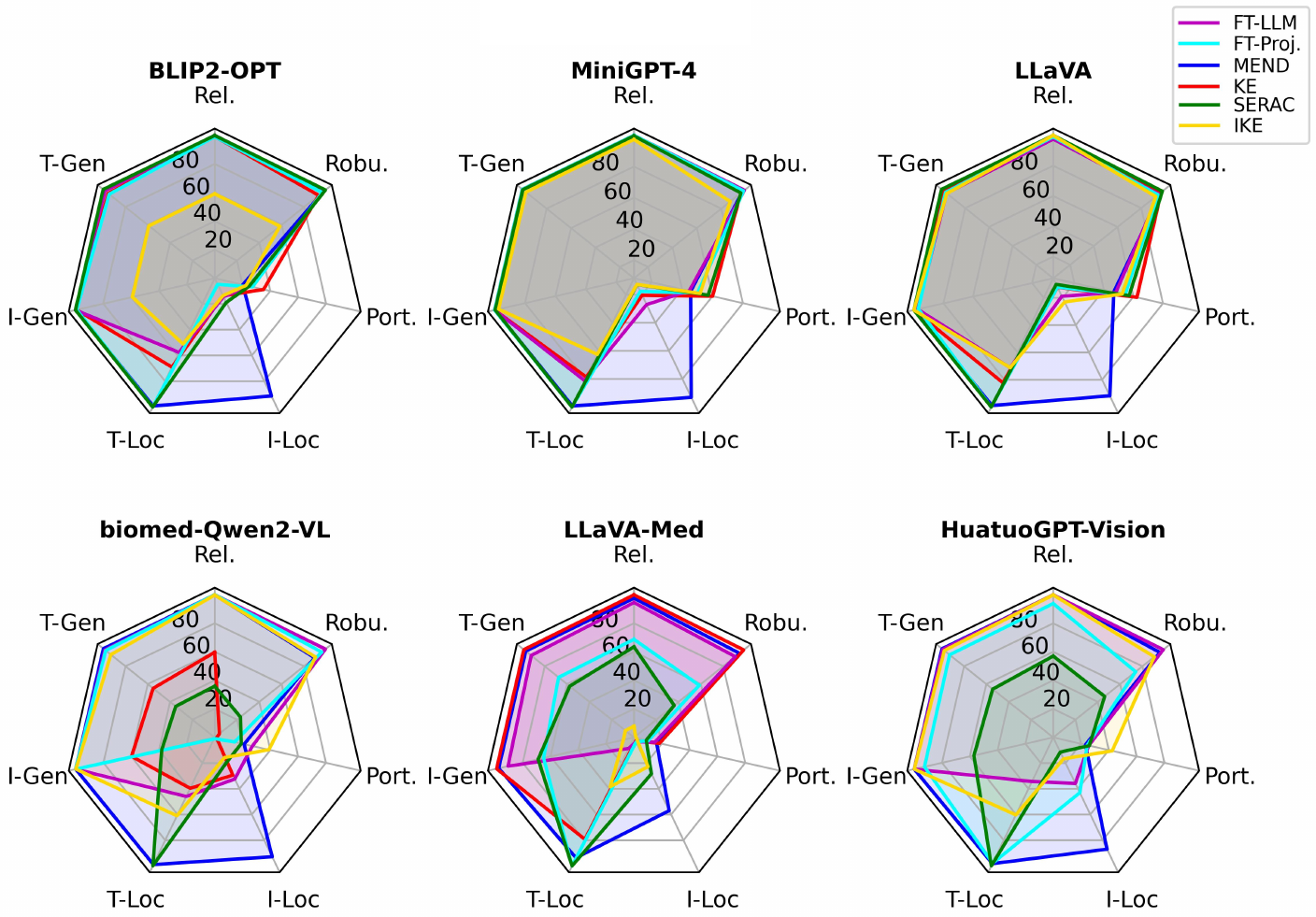}}
\caption{Comparison of different knowledge editing methods on general MLLMs and medical MLLMs.}
\label{img:intro}
\end{figure}

Unlike knowledge editing in general fields, medical knowledge editing is uniquely challenging, and its nature is reflected in high risk, multimodal complexity, and high professionalism~\cite{xu2024mlevlm}. First, medical knowledge often relies on the joint understanding of image evidence and text context~\cite{chen2025mimo}. Second, in clinical applications, incorrect knowledge editing may have serious consequences, so it is not only required to be accurate in facts, but also to comply with the latest medical guidelines and clinical reasoning standards~\cite{chen2025beyond}. Third, medical concepts usually have a hierarchical structure and are closely related to each other, and knowledge editing requires extremely high precision and granularity~\cite{huang2025medical}. These characteristics determine that medical knowledge editing requires more professional editing mechanisms and evaluation standards than general methods.

To address these challenges, we propose \textbf{MedMKEB}, the first comprehensive benchmark tailored for the task of knowledge editing in medical multimodal large language models. MedMKEB consists of high-quality medical visual question answering data covering multiple medical modalities, tasks, and body parts. It supports systematic evaluation from five fundamental perspectives: \textbf{(1) Reliability}: whether the model is consistent with newly injected knowledge; \textbf{(2) Locality}: whether irrelevant knowledge is unaffected; \textbf{(3) Generality}: whether the model can apply edited knowledge to semantically similar but unseen cases; \textbf{(4) Portability}: whether the knowledge can be transferred to related reasoning contexts. In addition, we propose the fifth dimension \textbf{(5) Robustness} to test the stability of the edited model under adversarial prompts commonly seen in clinical settings.

In the process of constructing MedMKEB, we designed a series of challenging editing samples, including counterfactual replacement, text semantic rephrasing, image replacement, and multi-hop reasoning chain construction based on a medical knowledge graph. To simulate the interference in the real clinical environment, we also designed a variety of adversarial question injection strategies, such as ambiguous wording, misleading context, redundant clinical information, etc. All generated data are manually reviewed by medical experts to ensure their professionalism and validity. Based on MedMKEB, we test state-of-the-art general and medical multimodal large language models. Extensive experiments have shown that existing knowledge editing methods exhibit limitations in the medical domain, underscoring the need for specialized editing strategies.

In conclusion, our contributions are as follows:
\begin{itemize}
    \item We propose MedMKEB, the first multimodal medical knowledge editing benchmark, covering diverse modalities, tasks, and expert-verified edits.
    \item We design a multidimensional evaluation framework that covers five critical aspects, including reliability, locality, generality, portability, and robustness, which provide a holistic view of editing performance in real-world clinical scenarios.
    \item We conduct large-scale experiments and reveal limitations of existing editing methods, underscoring the need for specialized medical approaches.
\end{itemize}

\section{Related Work}


\subsection{Knowledge Editing Methods}

Knowledge editing has emerged to address issues like knowledge lag, misinformation, and customization needs in Large Language Models (LLMs) by enabling quick and efficient modification or injection of specific knowledge without large-scale retraining. Early methods are mainly based on fine-tuning or parameter-efficient fine-tuning techniques to achieve knowledge modification through local parameter updates. However, these methods are usually costly and have catastrophic forgetting problems. To address these challenges, SERAC~\cite{mitchell2022memory} proposed a knowledge editing method that combines retrieval enhancement and counterfactual comparison, thus leading SERAC to be able to locate contextual information related to the target knowledge and optimize the pertinence and accuracy of editing. 
IKE~\cite{zheng2023can} emphasizes the realization of knowledge editing through context control without directly modifying model parameters, guiding the model to express updated knowledge during reasoning. Knowledge Editor~\cite{de2021editing} avoids catastrophic forgetting caused by the training process by training a hypernetwork. MEND~\cite{mitchell2022fast} proposes to achieve fast and low-cost knowledge editing by learning differentiable gradient transformations, and effectively avoids the impact on other irrelevant knowledge. ROME~\cite{meng2022locating} directly modifies model parameters through sparse low-rank matrix updates, injects new factual information into the model, and has stronger controllability. These methods have gradually promoted the development of knowledge editing technology for large language models.

\subsection{Knowledge Editing Benchmarks}

To assess the effectiveness and reliability of knowledge editing methods, researchers have developed a series of standardized evaluation benchmarks. The commonly used ZsRE~\cite{levy2017zero} dataset tests the accuracy and generative capacity of the model in knowledge editing through sentence-level factual questions and answers. The CounterFact~\cite{meng2022locating} benchmark focuses on anti-forgetting and editing scope control, evaluating whether the model retains the original knowledge system while editing new knowledge. In the multimodal field, the benchmark of knowledge editing has made initial progress. MMEdit~\cite{cheng2023can} is the first systematic knowledge editing evaluation framework for MLLMs. The benchmark covers a variety of multimodal tasks such as visual question answering and image caption generation. VLKEB~\cite{huang2024vlkeb} constructs further generalization problems by performing multi-hop retrieval in the knowledge graph to test the portability of the model after editing knowledge. MMKE-BENCH~\cite{du2025mmke} proposes a more challenging multimodal knowledge editing evaluation scheme, including Visual Semantic Editing and User-Specific Editing, which helps promote the practical verification of multimodal knowledge editing methods in complex and real environments. In the medical field, the current main work still focuses on single-modal knowledge editing, such as the counterfactual dataset MedCF~\cite{xu2024editing} and MedEditBench~\cite{chen2025beyond}, but there is still a lack of knowledge editing evaluation for medical MLLMs. Our work aims to fill this gap.
\section{Preliminary}

\subsection{Problem Definition}

Medical multimodal knowledge can be expressed as a triple $k=<i, x, a>$, where $i$ is an image, $x$ is a knowledge question related to the image, and $a$ is the answer to $x$. In the process of knowledge editing, we hope to modify the answer to obtain $k_e=<i, x,a_e>$, to correct the intrinsic knowledge of the model. Specifically, consider the basic edited dataset: $\mathcal{D}_{edit}=\left \{ (i, x, a, a_e)_j \right\}^{|\mathcal{D}|}_{j=1}$, each piece of data contains image $i$, question input $x$, original answer $a$, and edited answer $a_e$. The medical multimodal large language model $f_{M}$ can be expressed as: $f_{M}(i, x; \theta)=a$, where $\theta = \theta_{v}\times \theta_{llm}$, represents the original model parameters, $\theta_{v}$ is the visual module parameter, and $\theta_{llm}$ is the language model parameter. After knowledge editing, the model parameters are updated from $\theta$ to $\theta'$, i.e., $\theta' = \text{KE}(f_{M}(\theta))$. The expected output of the model for the original input also changes from $a$ to $a_e$:$f_{M}(i, x; \theta') = a_e$.

\begin{figure*}[!ht]
\centerline{\includegraphics[width=0.9\linewidth]{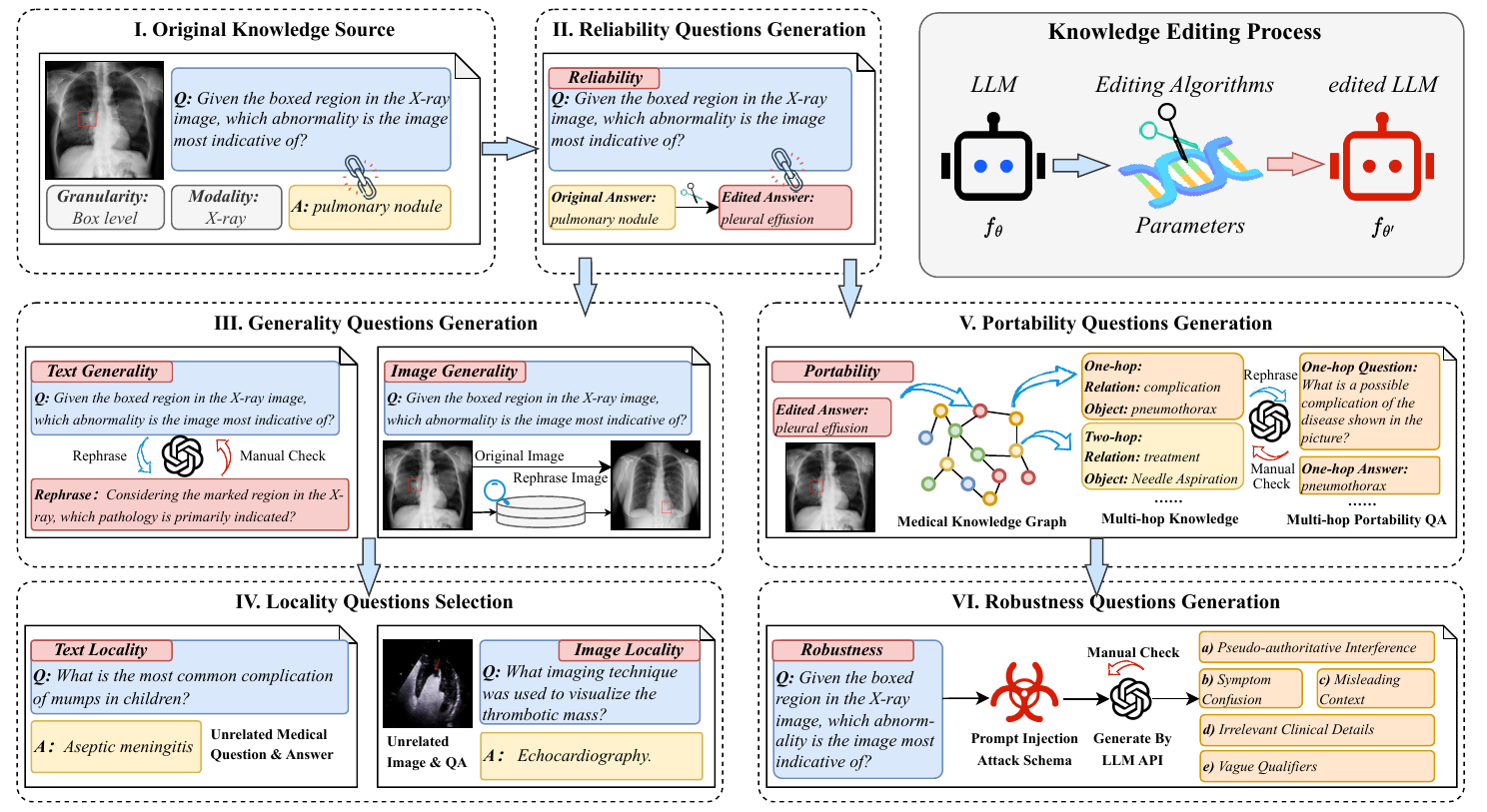}}
\caption{The construction pipeline of MedMKEB.}
\label{img: dialoguesample}
\end{figure*}

\subsection{Metrics}

In order to comprehensively evaluate the knowledge editing effect of medical multimodal large language models, this benchmark designed a systematic metric system from the dimensions of reliability, locality, generality, and portability. In addition, we propose the robustness evaluation metric in knowledge editing for the first time. The specific definitions of these metrics are as follows.

\paragraph{Reliability} is used to measure the effectiveness of knowledge editing operations on specific goals and ensure that the model output is consistent with the updated medical facts. For a given edit dataset $\mathcal{D}_{edit}$, we sample a quadruple $(i, x, a, a_e)$ from it, and the goal is to modify the original answer $a$ to the expected correct answer $a_e$. The reliability calculation formula is:
\begin{equation}
    \mathcal{M}_{rel}=\mathop{\mathbb{E}}\limits_{(i,x,a,a_e)\sim D_{edit}}[\mathds{1}\{f_M(i, x;\theta')=a_e\}],
\end{equation}
where $\mathds{1}\{\cdot\}$ is an indicator function that returns 1 if the output of the model matches the target answer.

\paragraph{Locality} is used to measure the impact of knowledge editing operations on other unedited knowledge and task capabilities of the model. Ideally, editing should be locally effective and have minimal perturbation to global knowledge. Specifically, for the two modalities of image and text, the locality is defined as follows:
\begin{gather}
    \mathcal{M}_{loc}^{txt}=\mathop{\mathbb{E}}\limits_{(x,a,a_e)\sim D_{loc}^{txt}}[\mathds{1}\{f_M(x;\theta')=f_M{(x;\theta)}\}],\\
    \mathcal{M}_{loc}^{img}=\mathop{\mathbb{E}}\limits_{(i,x,a,a_e)\sim D_{loc}^{img}}[\mathds{1}\{f_M(i,x;\theta')=f_M{(i,x;\theta)}\}],
\end{gather}
where $\mathcal{D}^{txt}_{loc}$ and $\mathcal{D}_{loc}^{img}$ are the text and visual question-answering dataset outside the distribution of the knowledge editing dataset $\mathcal{D}_{edit}$, respectively. 

\paragraph{Generality} is used to measure the generalization ability of the model to the new knowledge after editing, that is, whether the model can give answers that are consistent with medical facts in the adjacent semantic space that is not directly edited but related to the target knowledge. This metric reflects that knowledge editing is not limited to single-point corrections. The specific definition is as follows:
\begin{gather}
    \mathcal{M}_{gen}^{txt}=\mathop{\mathop{\mathbb{E}}\limits_{(i,x,a,a_e)\sim D_{edit}}}\limits_{x_g\in\mathcal{N}(x)}[\mathds{1}\{f_M(i,x_g;\theta')=a_e\}], \\
    \mathcal{M}_{gen}^{img}=\mathop{\mathop{\mathbb{E}}\limits_{(i,x,a,a_e)\sim D_{edit}}}\limits_{i_g\in\mathcal{N}(i)}[\mathds{1}\{f_M(i_g,x;\theta')=a_e\}],
\end{gather}
where $\mathcal{N}(x)$ and $\mathcal{N}(i)$ represent the neighborhood sets of the original input $x$ or $i$ in the semantic space in text and image modalities, respectively.

\paragraph{Portability} is used to measure the transferability of knowledge editing capabilities on a wider range of related tasks. It can evaluate whether the edited model can be effectively applied to other knowledge related to the edited knowledge.
\begin{equation}
    \mathcal{M}_{port}=\mathop{\mathop{\mathbb{E}}\limits_{(i,x,a,a_e)\sim D_{edit}}}\limits_{x_p, a_p \sim \mathcal{P}(i, x, a, a_e)}[\mathds{1}\{f_M(i,x_p;\theta')=a_p\}],
\end{equation}
where $\mathcal{P}(i,x,a,a_e)$ represents the portability domain related to the original data, such as multi-hop knowledge, relation reversal, task migration, etc.

\paragraph{Robustness} is used to measure the stability of the edited knowledge when the model is exposed to adversarial perturbations in prompts, such as misleading rephrasings, distractors, or prompt injections. A robust model should still output the correct, edited medical knowledge even when the question is subtly altered or perturbed. Given a perturbation function $\mathcal{A}(\cdot)$ that generates adversarial versions of the original input $(i, x)$, the robustness is defined as:
\begin{equation}
    \mathcal{M}_{port}=\mathop{\mathop{\mathbb{E}}\limits_{(i,x,a,a_e)\sim D_{edit}}}\limits_{i^{adv}, x^{adv} \sim \mathcal{A}(i, x)}[\mathds{1}\{f_M(i^{adv},x^{adv};\theta')=a_e\}],
\end{equation}
where $(i^{adv},x^{adv})$ denotes the adversarially perturbed image-text pair derived from the original input $(i,x)$, and $f_M$ is expected to preserve the edited knowledge output $a_e$ despite such perturbations.

\section{Benchmark Construction}

The overall construction pipeline of MedMKEB is shown in Figure 2, which includes original knowledge source collection, construction of various questions, and manual check.

\subsection{Original Knowledge Source}

Our proposed MedMKEB is built on the OmniMedVQA~\cite{hu2024omnimedvqa}, which provides a comprehensive collection of medical images with relevant clinical questions. OmniMedVQA is widely used in medical visual question answering, covering multiple medical fields such as radiology, pathology, and ophthalmology. The images, as well as the corresponding questions and answers, provide an ideal data source for developing a knowledge editing benchmark for medical MLLMs.

Specifically, we set rules to filter the modality, task type, department, and perception granularity of questions, avoiding imbalance between categories while ensuring the diversity of questions. In the end, we screened 6987 high-quality multimodal question-answer pairs, covering 16 visual question-answering tasks, 19 human body parts, and 16 types of medical modalities. These question-answer pairs will serve as the MedMKEB's initial knowledge source. The final dataset is denoted as $\mathcal{D}$. The statistics and distribution of the data source are shown in Appendix A.

\linespread{1.1}
\begin{table}[t]
\tabcolsep=2pt
\centering
\caption{Comparison with previous knowledge editing benchmarks.}
\begin{tabular}{lccccc}
\toprule
\textbf{Name} & \textbf{Multimodal} & \textbf{Port.}& \textbf{Medical} & \textbf{Robu.} & \textbf{Nums}\\
\midrule
ZsRE
& \XSolidBrush  & \XSolidBrush & \XSolidBrush & \XSolidBrush & 1037 \\
MMEdit
& \Checkmark  & \XSolidBrush & \XSolidBrush & \XSolidBrush & 8439 \\
VLKEB
& \Checkmark  & \Checkmark & \XSolidBrush & \XSolidBrush & 8174 \\
MMKE-Bench
& \Checkmark  & \Checkmark & \XSolidBrush & \XSolidBrush & 2940 \\
MedCF
& \XSolidBrush  & \XSolidBrush & \Checkmark & \XSolidBrush & 4025 \\
MedEditBench
& \XSolidBrush  & \XSolidBrush & \Checkmark & \XSolidBrush & 930 \\
\midrule
MedMKEB 
& \Checkmark  & \Checkmark & \Checkmark & \Checkmark & 6987 \\
\bottomrule
\end{tabular}
\label{tab:comparison_benchmarks}
\end{table}

\subsection{Evaluation Questions Generation}

To evaluate the performance of knowledge editing in medical multimodal models, we generate a set of evaluation questions that cover a variety of tasks, including reliability, generality, and locality. These questions are designed to test the model's ability to adapt to edited knowledge while maintaining high performance across different scenarios.

\linespread{1.1}
\begin{table*}[tbp!]
    \caption{Single editing results for general MLLMs. The best result and the second best result in each row are indicated by \textbf{bold} and \underline{underline}, respectively.} 
    \fontsize{8}{7}\selectfont    
    \centering
    \begin{tabular}{c|c|rrrrrrr}
    \toprule
    \multicolumn{1}{c|}{\textbf{Model}} 
    & \multicolumn{1}{c|}{\textbf{Editing Method}} 
    
    & \multicolumn{1}{c}{\textbf{Reliability}} 
    & \multicolumn{1}{c}{\textbf{T-Generality}} 
    & \multicolumn{1}{c}{\textbf{I-Generality}}
    & \multicolumn{1}{c}{\textbf{T-Locality}}
    & \multicolumn{1}{c}{\textbf{I-Locality}}
    & \multicolumn{1}{c}{\textbf{Portability}}
    & \multicolumn{1}{c}{\textbf{Robustness}} \\
    
    
    \midrule

    \multirow{7}{*}{BLIP2-OPT}
    & \multicolumn{1}{c|}{FT-LLM}
    
    & \multicolumn{1}{c}{100.0} 
    & \multicolumn{1}{c}{98.23} 
    & \multicolumn{1}{c}{99.98} 
    & \multicolumn{1}{c}{57.63}
    & \multicolumn{1}{c}{14.40}
    & \multicolumn{1}{c}{24.94}
    & \multicolumn{1}{c}{98.73} 
    \\

    & \multicolumn{1}{c|}{FT-Proj}
    
    & \multicolumn{1}{c}{99.09} 
    & \multicolumn{1}{c}{94.39} 
    & \multicolumn{1}{c}{99.02} 
    & \multicolumn{1}{c}{100.0}
    & \multicolumn{1}{c}{4.88}
    & \multicolumn{1}{c}{27.20}
    & \multicolumn{1}{c}{96.26}  \\
    \cmidrule{2-9}

    & \multicolumn{1}{c|}{IKE}
    
    & \multicolumn{1}{c}{59.12} 
    & \multicolumn{1}{c}{58.94} 
    & \multicolumn{1}{c}{59.15}
    & \multicolumn{1}{c}{51.11}
    & \multicolumn{1}{c}{13.54}
    & \multicolumn{1}{c}{22.53}
    & \multicolumn{1}{c}{58.44} \\

    & \multicolumn{1}{c|}{SERAC}
    
    & \multicolumn{1}{c}{\textbf{99.93}} 
    & \multicolumn{1}{c}{\textbf{99.77}} 
    & \multicolumn{1}{c}{\textbf{99.93}} 
    & \multicolumn{1}{c}{\textbf{100.0}}
    & \multicolumn{1}{c}{\underline{18.80}}
    & \multicolumn{1}{c}{\underline{24.92}}
    & \multicolumn{1}{c}{\textbf{99.10}} \\
    
    & \multicolumn{1}{c|}{MEND}
    
    & \multicolumn{1}{c}{\underline{99.82}}
    & \multicolumn{1}{c}{\underline{99.60}} 
    & \multicolumn{1}{c}{\underline{99.76}} 
    & \multicolumn{1}{c}{\underline{99.20}}
    & \multicolumn{1}{c}{\textbf{91.44}}
    & \multicolumn{1}{c}{20.27}
    & \multicolumn{1}{c}{\underline{97.45}} \\
    
    & \multicolumn{1}{c|}{KE}
    
    & \multicolumn{1}{c}{99.19} 
    & \multicolumn{1}{c}{95.29} 
    & \multicolumn{1}{c}{99.19} 
    & \multicolumn{1}{c}{69.05}
    & \multicolumn{1}{c}{13.75}
    & \multicolumn{1}{c}{\textbf{35.15}}
    & \multicolumn{1}{c}{93.09}
    \\
    \midrule

    \multirow{7}{*}{MiniGPT4}
    & \multicolumn{1}{c|}{FT-LLM}
    
    & \multicolumn{1}{c}{100.0} 
    & \multicolumn{1}{c}{99.87} 
    & \multicolumn{1}{c}{100.0} 
    & \multicolumn{1}{c}{81.22}
    & \multicolumn{1}{c}{27.03}
    & \multicolumn{1}{c}{44.71}
    & \multicolumn{1}{c}{98.97} 
    \\

    & \multicolumn{1}{c|}{FT-Proj}
    
    & \multicolumn{1}{c}{100.0} 
    & \multicolumn{1}{c}{99.85} 
    & \multicolumn{1}{c}{99.98} 
    & \multicolumn{1}{c}{100.0}
    & \multicolumn{1}{c}{17.67}
    & \multicolumn{1}{c}{50.23}
    & \multicolumn{1}{c}{98.22}  \\
    \cmidrule{2-9}

    & \multicolumn{1}{c|}{IKE}
    
    & \multicolumn{1}{c}{97.39} 
    & \multicolumn{1}{c}{96.96} 
    & \multicolumn{1}{c}{97.41} 
    & \multicolumn{1}{c}{62.51}
    & \multicolumn{1}{c}{12.70}
    & \multicolumn{1}{c}{51.84}
    & \multicolumn{1}{c}{87.32} \\

    & \multicolumn{1}{c|}{SERAC}
    
    & \multicolumn{1}{c}{99.65} 
    & \multicolumn{1}{c}{99.51} 
    & \multicolumn{1}{c}{99.65} 
    & \multicolumn{1}{c}{\textbf{99.97}}
    & \multicolumn{1}{c}{12.79}
    & \multicolumn{1}{c}{\underline{57.13}}
    & \multicolumn{1}{c}{\underline{96.09}} \\
    
    & \multicolumn{1}{c|}{MEND}
    
    & \multicolumn{1}{c}{\underline{99.91}} 
    & \multicolumn{1}{c}{\textbf{99.84}} 
    & \multicolumn{1}{c}{\underline{99.86}} 
    & \multicolumn{1}{c}{\underline{99.29}}
    & \multicolumn{1}{c}{\textbf{93.08}}
    & \multicolumn{1}{c}{45.69}
    & \multicolumn{1}{c}{\textbf{97.17}} \\
    
    & \multicolumn{1}{c|}{KE}
    
    & \multicolumn{1}{c}{\textbf{100.0}} 
    & \multicolumn{1}{c}{\underline{99.64}} 
    & \multicolumn{1}{c}{\textbf{99.96}} 
    & \multicolumn{1}{c}{78.71}
    & \multicolumn{1}{c}{\underline{20.41}}
    & \multicolumn{1}{c}{\textbf{60.05}}
    & \multicolumn{1}{c}{96.08}
    \\
    \midrule

    \multirow{7}{*}{LLaVA}
    & \multicolumn{1}{c|}{FT-LLM}
    
    & \multicolumn{1}{c}{97.54} 
    & \multicolumn{1}{c}{95.71} 
    & \multicolumn{1}{c}{97.52} 
    & \multicolumn{1}{c}{71.23}
    & \multicolumn{1}{c}{18.84}
    & \multicolumn{1}{c}{48.01}
    & \multicolumn{1}{c}{94.71} 
    \\

    & \multicolumn{1}{c|}{FT-Proj}
    
    & \multicolumn{1}{c}{99.07} 
    & \multicolumn{1}{c}{96.63} 
    & \multicolumn{1}{c}{97.02} 
    & \multicolumn{1}{c}{100.0}
    & \multicolumn{1}{c}{12.17}
    & \multicolumn{1}{c}{54.35}
    & \multicolumn{1}{c}{94.08}  \\
    \cmidrule{2-9}

    & \multicolumn{1}{c|}{IKE}
    
    & \multicolumn{1}{c}{\textbf{100.0}} 
    & \multicolumn{1}{c}{95.51} 
    & \multicolumn{1}{c}{\textbf{100.0}} 
    & \multicolumn{1}{c}{71.16}
    & \multicolumn{1}{c}{\underline{23.02}}
    & \multicolumn{1}{c}{52.35}
    & \multicolumn{1}{c}{92.25} \\

    & \multicolumn{1}{c|}{SERAC}
    
    & \multicolumn{1}{c}{\textbf{100.0}} 
    & \multicolumn{1}{c}{\textbf{99.90}} 
    & \multicolumn{1}{c}{\textbf{100.0}} 
    & \multicolumn{1}{c}{\textbf{99.98}}
    & \multicolumn{1}{c}{10.19}
    & \multicolumn{1}{c}{\underline{57.09}}
    & \multicolumn{1}{c}{\textbf{97.83}} \\
    
    & \multicolumn{1}{c|}{MEND}
    
    & \multicolumn{1}{c}{99.67} 
    & \multicolumn{1}{c}{\underline{99.60}} 
    & \multicolumn{1}{c}{99.69} 
    & \multicolumn{1}{c}{\underline{98.89}}
    & \multicolumn{1}{c}{\textbf{91.77}}
    & \multicolumn{1}{c}{46.66}
    & \multicolumn{1}{c}{96.79} \\
    
    & \multicolumn{1}{c|}{KE}
    
    & \multicolumn{1}{c}{99.96} 
    & \multicolumn{1}{c}{98.23} 
    & \multicolumn{1}{c}{99.96} 
    & \multicolumn{1}{c}{82.24}
    & \multicolumn{1}{c}{12.30}
    & \multicolumn{1}{c}{\textbf{62.35}}
    & \multicolumn{1}{c}{\underline{96.80}}
    \\
    
    \bottomrule
    \end{tabular}
    \label{tab:comparison1}
\end{table*}

\paragraph{Reliability Questions Generation} We construct counterfactual questions that challenge the model to deal with conflicting or newly updated medical knowledge. Counterfactual evaluation is currently used by a large number of knowledge editing benchmarks~\cite{huang2024vlkeb, xu2024editing, du2025mmke}, and these questions can evaluate whether the model can maintain its accuracy after the information is updated. Specifically, for each visual question-answer pair $(i, x, a)$ in $\mathcal{D}$, we replace the entity $a$ corresponding to the original answer with a different entity with the highest similarity in the options as the editing target $a_e$, and obtain the updated editing dataset $\mathcal{D}_{edit} = \left \{ (i, x, a, a_e)_j \right\}^{|\mathcal{D}|}_{j=1}$.

\paragraph{Generality Questions Generation} 
To test the generalization of edited knowledge, we construct new samples that preserve the underlying semantics while varying surface forms. For the textual modality, we apply the LLM API to rephrase the original questions $x$ into semantically equivalent variants $x_g$, ensuring that the core knowledge remains unchanged. For the visual modality, we identify and randomly replace the original image $i$ with a new image $i_g$ from the dataset that shares the same medical entity. This forms generalized datasets $\mathcal{D}_{gen}^{txt} = \left\{ (i, x_g, a) \right\}$ and $\mathcal{D}_{gen}^{img} = \left\{ (i_g, x, a) \right\}$, where $x_g \in \mathcal{N}(x)$ and $i_g \in \mathcal{N}(i)$, used to evaluate whether the edited knowledge can be transferred and generalized across similar but not identical contexts.

\paragraph{Locality Questions Selection} 
To assess the model’s ability to localize and isolate the effects of knowledge editing, we construct locality-sensitive evaluation sets. For the textual modality, we leverage the MedMCQA dataset~\cite{pal2022medmcqa} to identify questions that share similar structures and domains with $\mathcal{D}$ but are not affected by the editing operation. For the visual modality, we use PMC-VQA~\cite{zhang2023pmc} to sample medically related but semantically independent image-question pairs. This helps verify whether the knowledge editing operation preserves unrelated knowledge and avoids unintended side effects, forming a locality validation set $\mathcal{D}_{loc}^{txt} = \left\{ (x', a') \right\}$ and $\mathcal{D}_{loc}^{img} = \left\{ (i'', x'', a'') \right\}$.

\subsection{Portability and Robustness}
We further assess the portability and robustness of edited knowledge in complex medical reasoning and adversarial contexts. These two aspects help to ensure that the modified knowledge is not only effective in isolated cases but can be transferred and defended across more realistic scenarios.

\paragraph{Portability Questions Generation} Portability refers to the ability of a model to apply edited knowledge beyond the direct question where editing occurred~\cite{huang2024vlkeb}, to related contexts within a transferable scope. Formally, we denote this as testing whether the updated knowledge tuple $(i, x, a, a_e)$ can be generalized to a broader reasoning scope $\mathcal{P}(i, x, a, a_e)$. To build the one-hop portability dataset, we identify connected triples of the edited answer entity $a_e$, such that the model must reason across a single related fact. Suppose the editing operation modifies a sample $(i, x, a)$ into $(i, x, a_e)$, and there exists a fact $(a_e, r, o)$ in the medical knowledge graph. We then construct one-hop reasoning questions $q(o; r)$, which assess whether the model can incorporate $a_e$ in a semantically adjacent question. All such question-answer pairs are collected to form the one-hop portability evaluation dataset $\mathcal{D}_{port}^{(1)}$:
\begin{equation}
    \mathcal{D}^{(1)}_{port} = \{q(o;r)|(a_{e}, r, o)\in \mathcal{K}\},
\end{equation}
where $\mathcal{K}$ is the medical knowledge graph and we choose LMKG~\cite{yang2024lmkg} as the reference knowledge graph.

In multi-hop scenarios, evaluating portability becomes more challenging, as the model must integrate the edited knowledge through a chain of reasoning steps. We construct paths $\left\langle (i, x, a \rightarrow a_e), (s_1, r_1, o_1), \ldots, (s_n, r_n, o_n) \right\rangle$, where $a_e = s_1$ and $o_i = s_{i+1}$ for $i = 1$ to $n-1$. These paths form a compositional reasoning chain from the editing fact. Similarly, we get the multi-hop knowledge editing dataset: 
\begin{equation}
    \mathcal{D}^{(n)}_{port} = \{q(o;r_1, r_2, \dots, r_n) | \mathcal{C}\},
\end{equation}
where $\mathcal{C}$ is the multi-hop sequence constructed from the knowledge graph. This design allows us to assess whether the model can propagate the edited knowledge across complex medical reasoning chains, a crucial capability for real-world decision support in clinical settings.

\paragraph{Robustness Questions Generation} 
To test the robustness of knowledge editing, we chose the most common attack method in large language model security, i.e., prompt injection attack~\cite{liu2023prompt, liu2024automatic, clusmann2025prompt}. We designed a set of prompt-based perturbation schemes to simulate real attack problems in clinical scenarios~\cite{huang2025medical}. We apply five types of prompt injection attacks, including \textit{ 1) pseudo-authoritative interference, 2) symptom confusion, 3) misleading context, 4) irrelevant clinical details, and 5) vague characteristics}. Specific examples of attack content are shown in Appendix A. For each question $x$ in $\mathcal{D}_{edit}$, we use the large language model to generate the corresponding adversarial sample $x^{\text{adv}}$ and manually verify the generated questions:
\begin{equation}
    \mathcal{D}_{adv} = \{(i, x^{adv}, a_e) | x^{adv} \in \mathcal{A}(x)\},
\end{equation}
where $\mathcal{A}(x)$ is the scope of the perturbed problems generated based on the original question $x$. 

\linespread{1.1}
\begin{table*}[tbp!]
    \caption{Single editing results for medical MLLMs. The best result and the second best result in each row are indicated by \textbf{bold} and \underline{underline}, respectively.} 
    \fontsize{8}{7}\selectfont    
    \centering
    \begin{tabular}{c|c|rrrrrrr}
    \toprule
    \multicolumn{1}{c|}{\textbf{Model}} 
    & \multicolumn{1}{c|}{\textbf{Editing Method}} 
    
    & \multicolumn{1}{c}{\textbf{Reliability}} 
    & \multicolumn{1}{c}{\textbf{T-Generality}} 
    & \multicolumn{1}{c}{\textbf{I-Generality}}
    & \multicolumn{1}{c}{\textbf{T-Locality}}
    & \multicolumn{1}{c}{\textbf{I-Locality}}
    & \multicolumn{1}{c}{\textbf{Portability}}
    & \multicolumn{1}{c}{\textbf{Robustness}} \\
    
    
    \midrule

    \multirow{7}{*}{Biomed-Qwen2-VL}
    & \multicolumn{1}{c|}{FT-LLM}
    
    & \multicolumn{1}{c}{100.0} 
    & \multicolumn{1}{c}{98.37} 
    & \multicolumn{1}{c}{100.0} 
    & \multicolumn{1}{c}{45.93}
    & \multicolumn{1}{c}{32.37}
    & \multicolumn{1}{c}{27.12}
    & \multicolumn{1}{c}{99.29} 
    \\

    & \multicolumn{1}{c|}{FT-Proj}
    
    & \multicolumn{1}{c}{100.0} 
    & \multicolumn{1}{c}{97.73} 
    & \multicolumn{1}{c}{100.0} 
    & \multicolumn{1}{c}{1.18}
    & \multicolumn{1}{c}{1.01}
    & \multicolumn{1}{c}{14.33}
    & \multicolumn{1}{c}{95.97}  \\
    \cmidrule{2-9}

    & \multicolumn{1}{c|}{IKE}
    
    & \multicolumn{1}{c}{\underline{99.82}}
    & \multicolumn{1}{c}{\underline{93.11}} 
    & \multicolumn{1}{c}{\underline{99.82}} 
    & \multicolumn{1}{c}{60.98}
    & \multicolumn{1}{c}{15.69}
    & \multicolumn{1}{c}{\textbf{38.95}}
    & \multicolumn{1}{c}{\textbf{90.36}} \\

    & \multicolumn{1}{c|}{SERAC}
    
    & \multicolumn{1}{c}{35.97} 
    & \multicolumn{1}{c}{34.95} 
    & \multicolumn{1}{c}{38.02} 
    & \multicolumn{1}{c}{\textbf{99.99}}
    & \multicolumn{1}{c}{19.33}
    & \multicolumn{1}{c}{19.17}
    & \multicolumn{1}{c}{23.03} \\
    
    & \multicolumn{1}{c|}{MEND}
    
    & \multicolumn{1}{c}{\textbf{99.93}} 
    & \multicolumn{1}{c}{\textbf{99.78}} 
    & \multicolumn{1}{c}{\textbf{99.93}} 
    & \multicolumn{1}{c}{\underline{98.98}}
    & \multicolumn{1}{c}{\textbf{92.84}}
    & \multicolumn{1}{c}{\underline{21.16}}
    & \multicolumn{1}{c}{\underline{89.55}} \\
    
    & \multicolumn{1}{c|}{KE}
    
    & \multicolumn{1}{c}{59.68} 
    & \multicolumn{1}{c}{55.20} 
    & \multicolumn{1}{c}{59.77} 
    & \multicolumn{1}{c}{39.60}
    & \multicolumn{1}{c}{\underline{29.37}}
    & \multicolumn{1}{c}{1.32}
    & \multicolumn{1}{c}{4.17}
    \\
    \midrule
    
    \multirow{7}{*}{LLaVA-Med}
    & \multicolumn{1}{c|}{FT-LLM}
    
    & \multicolumn{1}{c}{94.53} 
    & \multicolumn{1}{c}{92.11} 
    & \multicolumn{1}{c}{90.5} 
    & \multicolumn{1}{c}{8.58}
    & \multicolumn{1}{c}{3.36}
    & \multicolumn{1}{c}{15.01}
    & \multicolumn{1}{c}{90.8} 
    \\

    & \multicolumn{1}{c|}{FT-Proj}
    
    & \multicolumn{1}{c}{68.74} 
    & \multicolumn{1}{c}{67.53} 
    & \multicolumn{1}{c}{64.96} 
    & \multicolumn{1}{c}{100.0}
    & \multicolumn{1}{c}{3.48}
    & \multicolumn{1}{c}{10.19}
    & \multicolumn{1}{c}{58.58}  \\
    \cmidrule{2-9}

    & \multicolumn{1}{c|}{IKE}
    
    & \multicolumn{1}{c}{7.86} 
    & \multicolumn{1}{c}{7.69} 
    & \multicolumn{1}{c}{7.61} 
    & \multicolumn{1}{c}{38.36}
    & \multicolumn{1}{c}{22.76}
    & \multicolumn{1}{c}{1.71}
    & \multicolumn{1}{c}{1.48} \\

    & \multicolumn{1}{c|}{SERAC}
    
    & \multicolumn{1}{c}{63.49} 
    & \multicolumn{1}{c}{57.57} 
    & \multicolumn{1}{c}{68.85} 
    & \multicolumn{1}{c}{\textbf{100.0}}
    & \multicolumn{1}{c}{\underline{28.27}}
    & \multicolumn{1}{c}{8.82}
    & \multicolumn{1}{c}{36.02} \\
    
    & \multicolumn{1}{c|}{MEND}
    
    & \multicolumn{1}{c}{\underline{97.59}}
    & \multicolumn{1}{c}{\underline{97.02}}
    & \multicolumn{1}{c}{\underline{97.22}}
    & \multicolumn{1}{c}{\underline{93.67}}
    & \multicolumn{1}{c}{\textbf{56.95}}
    & \multicolumn{1}{c}{\underline{16.22}}
    & \multicolumn{1}{c}{\underline{94.58}} \\
    
    & \multicolumn{1}{c|}{KE}
    
    & \multicolumn{1}{c}{\textbf{99.91}} 
    & \multicolumn{1}{c}{\textbf{98.67}} 
    & \multicolumn{1}{c}{\textbf{98.73}} 
    & \multicolumn{1}{c}{79.29}
    & \multicolumn{1}{c}{2.79}
    & \multicolumn{1}{c}{\textbf{18.39}}
    & \multicolumn{1}{c}{\textbf{98.35}}
    \\
    \midrule

    \multirow{6}{*}{HuatuoGPT-Vision}
    & \multicolumn{1}{c|}{FT-LLM}
    
    & \multicolumn{1}{c}{100.0} 
    & \multicolumn{1}{c}{99.02} 
    & \multicolumn{1}{c}{100.0} 
    & \multicolumn{1}{c}{34.35}
    & \multicolumn{1}{c}{35.78}
    & \multicolumn{1}{c}{25.56}
    & \multicolumn{1}{c}{98.69} 
    \\

    & \multicolumn{1}{c|}{FT-Proj}
    
    & \multicolumn{1}{c}{93.89} 
    & \multicolumn{1}{c}{92.94} 
    & \multicolumn{1}{c}{92.60} 
    & \multicolumn{1}{c}{97.89}
    & \multicolumn{1}{c}{43.03}
    & \multicolumn{1}{c}{24.51}
    & \multicolumn{1}{c}{73.94}  \\
    \cmidrule{2-9}

    & \multicolumn{1}{c|}{IKE}
    
    & \multicolumn{1}{c}{\underline{99.95}} 
    & \multicolumn{1}{c}{\underline{97.68}} 
    & \multicolumn{1}{c}{\underline{99.95}} 
    & \multicolumn{1}{c}{60.33}
    & \multicolumn{1}{c}{\underline{16.68}}
    & \multicolumn{1}{c}{\textbf{42.43}}
    & \multicolumn{1}{c}{\underline{90.68}} \\

    & \multicolumn{1}{c|}{SERAC}
    
    & \multicolumn{1}{c}{57.05} 
    & \multicolumn{1}{c}{54.32} 
    & \multicolumn{1}{c}{57.02} 
    & \multicolumn{1}{c}{\textbf{100.0}}
    & \multicolumn{1}{c}{11.45}
    & \multicolumn{1}{c}{\underline{25.95}}
    & \multicolumn{1}{c}{46.15} \\
    
    & \multicolumn{1}{c|}{MEND}
    
    & \multicolumn{1}{c}{\textbf{99.97}} 
    & \multicolumn{1}{c}{\textbf{99.72}} 
    & \multicolumn{1}{c}{\textbf{99.99}} 
    & \multicolumn{1}{c}{\underline{98.29}}
    & \multicolumn{1}{c}{\textbf{86.99}}
    & \multicolumn{1}{c}{23.20}
    & \multicolumn{1}{c}{\textbf{95.64}}
    \\
    \bottomrule
    \end{tabular}
    \label{tab:comparison2}
\end{table*}

\subsection{Human Check and Statistics}\label{sec:HumanCheck}

To ensure the accuracy and consistency of the benchmark, human checks are incorporated into each stage of the construction process. To validate the quality of the answers provided by the model, three medical experts manually review the entire set of questions and correct unreasonable questions. These checks help identify any discrepancies or potential errors in model reasoning and knowledge application.

\paragraph{Statistics}
MedMKEB contains 6987 knowledge question answering pairs and 13060 images. The knowledge covers 16 medical vqa tasks and includes 2688 portability questions up to 3-hop and 721 robustness questions. We divide the dataset into training and validation sets in a ratio of 6:4. More details of MedMKEB can be found in Appendix A. The comparison with other knowledge editing benchmarks is shown in Table 1. Our proposed MedMKEB is the first comprehensive knowledge editing benchmark for medical multimodal large language models.

\section{Experiments}

We conducted single and sequential knowledge editing experiments on six MLLMs with different parameter sizes and architectures, including three general models: BLIP-OPT~\cite{li2023blip}, MiniGPT-4~\cite{zhuminigpt}, and LLaVA~\cite{liu2023visual}, and three medical models: LLaVA-Med~\cite{li2023llava}, Biomed-Qwen2-VL~\cite{cheng2024domain}, and HuatuoGPT-Vision~\cite{chen2024huatuogpt}. More details are shown in Appendix B.

\subsection{Editing Methods and Experimental Settings}

\paragraph{Methods} Following previous studies, we selected five representative knowledge editing methods as comparison methods, including fine-tuning, Knowledge Editor (KE)~\cite{de2021editing}, MEND~\cite{mitchell2022fast}, SERAC~\cite{mitchell2022memory}, and IKE~\cite{zheng2023can}. Fine-tuning includes fine-tuning of the LLMs (FT-LLM) or the visual language alignment module (FT-Proj). To avoid catastrophic forgetting, we only fine-tune the last layer of the large language model. Detailed introductions to these knowledge editing algorithms are shown in Appendix B.

\paragraph{Settings} We use the EasyEdit framework~\cite{wang2024easyedit} and complete the support for medical multimodal large language models. All experiments are performed on one NVIDIA A800 GPU with 80GB. For each model, we use its public pre-trained weights. The batch size of all experiments is uniformly set to 1, the text editing loss weight and image editing loss weight are set to 0.1, and the locality loss weight is set to 1. See Appendix C for more details. 

\subsection{Single Editing Results and Analysis}

\begin{figure*}[!t]
\centerline{\includegraphics[width=0.85\linewidth]{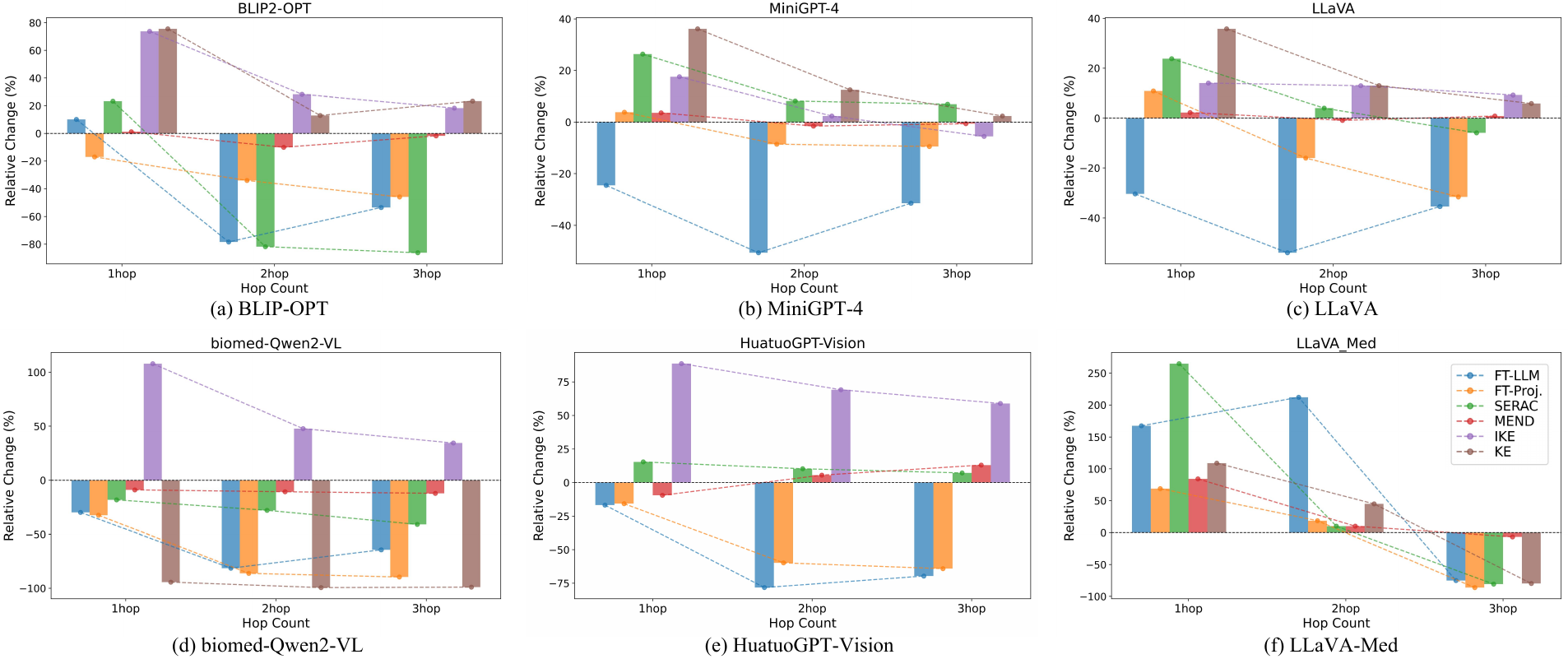}}
\caption{The relative change of multi-hop portability. $Relative  Change (\%) = \frac{(Portability-Base Portability)}{(Base Portability)}$.}
\label{img:multihop}
\end{figure*}

\begin{figure*}[!t]
\centerline{\includegraphics[width=0.85\linewidth]{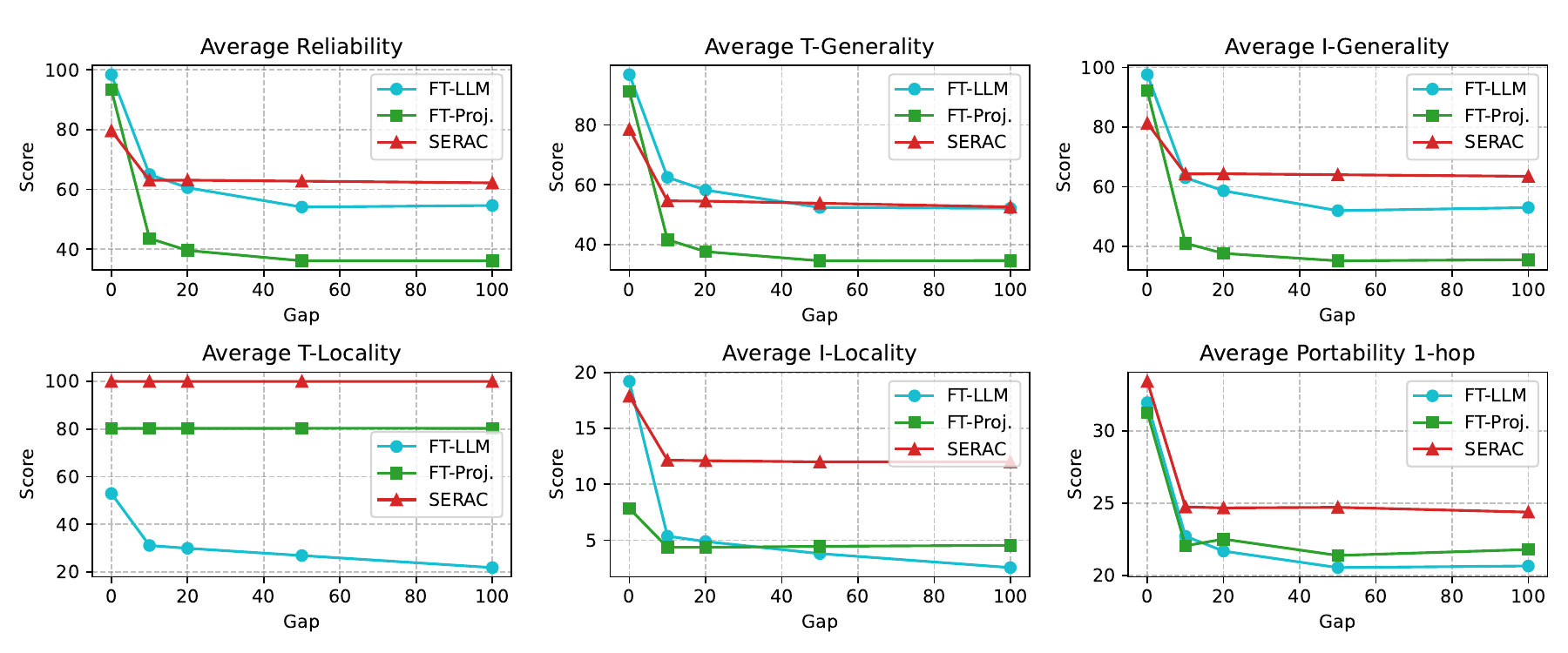}}
\caption{Average results in sequential editing.}
\label{img:sequential}
\end{figure*}

We first conducted single editing experiments. Table 2 and Table 3 show the results for general models and medical models, respectively. The visual comparison results of different methods are shown in Figure 1. From the observations, we can draw the following conclusions. 1) For general models, the reliability of various algorithms is very close, and in most cases it can reach more than 99\%, indicating that these algorithms can successfully modify the edited knowledge. For medical models, the reliability of the SERAC algorithm decreased significantly, and none of them reached 70\%, which is related to the ability of the counterfactual model. 2) Compared to the fine-tuning method, on different models, several carefully designed knowledge editing algorithms can have higher locality while ensuring generalization. It shows that knowledge editing algorithms can effectively avoid the influence of irrelevant knowledge. The MEND algorithm has a higher I-Locality score than other algorithms, whether it is a general model or a medical model. 3) For general models and LLaVA-Med, KE obtained the best portability results. This shows that hypernetwork-based editing methods can more effectively cope with this challenge. In addition, all knowledge editing methods have good generalization, but have difficulties in portability. This highlights the difficulty in applying edited knowledge to new content. 4) Regarding robustness, existing knowledge editing methods show a decline in both general models and medical models. However, FT-LLM hardly shows a decline in robustness, which indicates that the existing knowledge editing algorithms lack defense against prompt injection attacks.

\linespread{1.1}
\begin{table*}[htbp!]
    \caption{The time and memory cost of the knowledge editing algorithm on different models. The time cost represents the average editing time of each knowledge item (s/item), and the memory cost represents the memory size (GB) occupied when training or testing the model. The KE algorithm is out-of-memory on HuatuoGPT-Vision.} 

    \fontsize{8}{10}\selectfont    
    \centering
    \begin{tabular}{c|rr|rr|rr|rr|rr|rr}
    \toprule
    \multirow{2}{*}{\textbf{Model}} 
    & \multicolumn{2}{c|}{\textbf{FT-LLM}}
    & \multicolumn{2}{c|}{\textbf{FT-Proj}}
    & \multicolumn{2}{c|}{\textbf{IKE}}
    & \multicolumn{2}{c|}{\textbf{SERAC}}
    & \multicolumn{2}{c|}{\textbf{MEND}}
    & \multicolumn{2}{c}{\textbf{KE}}\\
    \cline{2-13}

    & \multicolumn{1}{c}{Time} 
    & \multicolumn{1}{c|}{Memory}
    & \multicolumn{1}{c}{Time} 
    & \multicolumn{1}{c|}{Memory}
    & \multicolumn{1}{c}{Time} 
    & \multicolumn{1}{c|}{Memory}
    & \multicolumn{1}{c}{Time} 
    & \multicolumn{1}{c|}{Memory}
    & \multicolumn{1}{c}{Time} 
    & \multicolumn{1}{c|}{Memory}
    & \multicolumn{1}{c}{Time} 
    & \multicolumn{1}{c}{Memory}\\
    \midrule

    \multicolumn{1}{l|}{BLIP2-OPT} 
    
    & \multicolumn{1}{c}{2.61} 
    & \multicolumn{1}{c|}{20.92}
    & \multicolumn{1}{c}{3.82}
    & \multicolumn{1}{c|}{21.45}

    & \multicolumn{1}{c}{2.90} 
    & \multicolumn{1}{c|}{13.98}
    & \multicolumn{1}{c}{7.78}
    & \multicolumn{1}{c|}{14.76}
    & \multicolumn{1}{c}{11.09} 
    & \multicolumn{1}{c|}{21.07}
    & \multicolumn{1}{c}{1.95}
    & \multicolumn{1}{c}{34.98}\\

    \multicolumn{1}{l|}{MiniGPT4} 
    
    & \multicolumn{1}{c}{3.06} 
    & \multicolumn{1}{c|}{45.37}
    & \multicolumn{1}{c}{3.88}
    & \multicolumn{1}{c|}{45.75}

    & \multicolumn{1}{c}{2.88} 
    & \multicolumn{1}{c|}{22.04}
    & \multicolumn{1}{c}{13.26}
    & \multicolumn{1}{c|}{52.48}
    & \multicolumn{1}{c}{9.91} 
    & \multicolumn{1}{c|}{39.85}
    & \multicolumn{1}{c}{2.32}
    & \multicolumn{1}{c}{60.15}\\

    \multicolumn{1}{l|}{LLaVA} 
    
    & \multicolumn{1}{c}{10.80} 
    & \multicolumn{1}{c|}{35.09}
    & \multicolumn{1}{c}{16.11}
    & \multicolumn{1}{c|}{35.52}

    & \multicolumn{1}{c}{2.20} 
    & \multicolumn{1}{c|}{22.41}
    & \multicolumn{1}{c}{27.54}
    & \multicolumn{1}{c|}{77.05}
    & \multicolumn{1}{c}{12.31} 
    & \multicolumn{1}{c|}{49.87}
    & \multicolumn{1}{c}{2.54}
    & \multicolumn{1}{c}{51.40}\\

    \multicolumn{1}{l|}{Biomed-Qwen2-VL} 
    
    & \multicolumn{1}{c}{2.14} 
    & \multicolumn{1}{c|}{12.41}
    & \multicolumn{1}{c}{2.55}
    & \multicolumn{1}{c|}{12.41}

    & \multicolumn{1}{c}{2.48} 
    & \multicolumn{1}{c|}{9.78}
    & \multicolumn{1}{c}{14.14}
    & \multicolumn{1}{c|}{28.22}
    & \multicolumn{1}{c}{6.97} 
    & \multicolumn{1}{c|}{18.75}
    & \multicolumn{1}{c}{1.34}
    & \multicolumn{1}{c}{31.01}\\

    \multicolumn{1}{l|}{LLaVA-Med} 
    
    & \multicolumn{1}{c}{12.64} 
    & \multicolumn{1}{c|}{52.07}
    & \multicolumn{1}{c}{18.70}
    & \multicolumn{1}{c|}{47.99}

    & \multicolumn{1}{c}{2.96} 
    & \multicolumn{1}{c|}{22.45}
    & \multicolumn{1}{c}{27.20}
    & \multicolumn{1}{c|}{75.40}
    & \multicolumn{1}{c}{13.67} 
    & \multicolumn{1}{c|}{55.49}
    & \multicolumn{1}{c}{3.70}
    & \multicolumn{1}{c}{61.52}\\

    \multicolumn{1}{l|}{HuatuoGPT-Vision} 
    
    & \multicolumn{1}{c}{16.22} 
    & \multicolumn{1}{c|}{46.60}
    & \multicolumn{1}{c}{25.76}
    & \multicolumn{1}{c|}{59.34}

    & \multicolumn{1}{c}{1.81} 
    & \multicolumn{1}{c|}{34.96}
    & \multicolumn{1}{c}{43.33}
    & \multicolumn{1}{c|}{79.23}
    & \multicolumn{1}{c}{9.89} 
    & \multicolumn{1}{c|}{57.26}
    & \multicolumn{1}{c}{/}
    & \multicolumn{1}{c}{OOM}\\

    \bottomrule
    \end{tabular}
    \label{tab:ablation}
\end{table*}

Furthermore, by comparing the results in Table 2 and Table 3, we can draw the following conclusions: 1) The IKE algorithm works well on LLaVA but poorly on LLaVA-Med, which may be due to the loss of contextual learning ability after fine-tuning of LLaVA-Med, because IKE relies on the model's own contextual ability. 2) The SERAC algorithm generally performs well in general models, but its reliability in medical models decreases significantly. This shows. 3) In medical models, the reliability and portability of all algorithms have decreased. This shows that in the knowledge editing task of medical multimodal large language models, algorithms need higher accuracy and adaptability to specific domain knowledge. 4) Overall, the MEND algorithm has the best knowledge editing evaluation indicators, higher multimodal locality advantages, and is suitable for both general and medical models. 

Existing medical model knowledge editing algorithms still need significant improvement. 1) Model aspect. Medical MLLMs are often obtained by direct fine-tuning from general MLLMs, which overfit to the training data, resulting in reduced generality when facing unseen data. At the same time, the model's context learning ability is weakened, and knowledge editing algorithms such as IKE and SERAC, whose parameters are not updated, cannot answer questions correctly. 2) Algorithm aspect. The existing knowledge editing algorithms for MLLMs mainly focus on LLM parameters and lack joint optimization of visual modules and text modules, resulting in poor performance of the algorithm's image locality evaluation indicators. Among these algorithms, meta-learning-based algorithms such as KE and MEND perform well, but are still not as good as those in general models. More analyses are listed in Appendix D. These conclusions show that there is still a need to design specific knowledge editing algorithms for medical MLLMs.

\subsection{Multi-hop Portability Results}

To comprehensively evaluate portability, we show the relative change in portability performance over multiple hops in Figure 3. In general, the relative performance of almost all models and methods shows a downward trend with the increase of the number of hops, indicating that knowledge transfer faces greater challenges in multi-hop reasoning paths. In terms of model dimension, LLaVA-Med performs best in the 1-hop scenario, but this advantage rapidly weakens as the number of hops increases, and even turns to negative growth in the 3-hop scenario, indicating that the model is extremely sensitive to local knowledge updates but unstable in long-distance transmission. The IKE method shows strong stability in multiple models, and its performance degradation is significantly smaller than that of other methods. In addition, SERAC and MEND show good anti-degradation capabilities in some models, especially in the 2-hop and 3-hop scenarios of MiniGPT-4 and HuatuoGPT-Vision, which can still maintain relatively stable performance. The FT-LLM method shows drastic fluctuations in multi-hop tasks, indicating that it is difficult to support the deep transmission of complex knowledge by only adjusting the parameters of the last layer of LLM.

\subsection{Sequential Editing Results and Analysis}

We also conducted experiments on sequential editing, and the average results of the 3 algorithms are shown in Figure 4. We set the sequence editing gaps to 10, 20, 50, and 100, and the details are shown in Appendix C. The experimental results show that SERAC shows good stability in all metrics, especially when dealing with large gaps, its performance is better than FT. As the gap increases, the reliability of FT decreases, and the locality decreases significantly. This shows that the effect of sequential editing needs to be improved. More experiments and analysis are shown in Appendix D.

\subsection{Editing Cost Analysis}

To better measure the efficiency differences between knowledge editing algorithms, we calculated their editing efficiency on different models, including time and memory usage. The experimental results are shown in Table 4. Because the IKE algorithm does not require additional training, it has the lowest usage of time and memory. Among algorithms that require updating model parameters, SERAC introduces an additional counterfactual model, which results in higher memory usage and time overhead. In contrast, the MEND algorithm has a lower time cost, similar to fine-tuning. The KE algorithm showed a higher memory cost, and the HuatuoGPT-Vision model experienced an out-of-memory error (OOM), which prevented it from training successfully.
\section{Conclusion and Future Work}

In this paper, we propose MedMKEB, the first benchmark for evaluating knowledge editing in multimodal large language models for medicine. MedMKEB evaluates five key aspects: reliability, locality, generality, portability, and robustness. Extensive experiments show that current editing methods struggle to handle the high precision and multimodal nature of medical knowledge. MedMKEB fills a critical gap and lays the foundation for developing safer and more effective knowledge editing techniques in medicine. In the future, more types of attack questions will be added to further evaluate the robustness of algorithms.

\bibliography{submission}

\begin{thebibliography}{42}
\providecommand{\natexlab}[1]{#1}

\bibitem[{Beurer-Kellner et~al.(2025)Beurer-Kellner, Cre{\c{t}}u, Debenedetti, Dobos, Fabian, Fischer, Froelicher, Grosse, Naeff, Ozoani et~al.}]{beurer2025design}
Beurer-Kellner, L.; Cre{\c{t}}u, B. B. A.-M.; Debenedetti, E.; Dobos, D.; Fabian, D.; Fischer, M.; Froelicher, D.; Grosse, K.; Naeff, D.; Ozoani, E.; et~al. 2025.
\newblock Design Patterns for Securing LLM Agents against Prompt Injections.
\newblock \emph{arXiv preprint arXiv:2506.08837}.

\bibitem[{Chen et~al.(2024)Chen, Gui, Ouyang, Gao, Chen, Chen, Wang, Zhang, Cai, Ji et~al.}]{chen2024huatuogpt}
Chen, J.; Gui, C.; Ouyang, R.; Gao, A.; Chen, S.; Chen, G.~H.; Wang, X.; Zhang, R.; Cai, Z.; Ji, K.; et~al. 2024.
\newblock Huatuogpt-vision, towards injecting medical visual knowledge into multimodal llms at scale.
\newblock \emph{arXiv preprint arXiv:2406.19280}.

\bibitem[{Chen et~al.(2025{\natexlab{a}})Chen, Luo, Qiu, Cao, Yang, and Pan}]{chen2025beyond}
Chen, S.; Luo, L.; Qiu, Z.; Cao, Y.; Yang, C.; and Pan, S. 2025{\natexlab{a}}.
\newblock Beyond Memorization: A Rigorous Evaluation Framework for Medical Knowledge Editing.
\newblock \emph{arXiv preprint arXiv:2506.03490}.

\bibitem[{Chen et~al.(2025{\natexlab{b}})Chen, Xu, Huang, Zhan, Wang, Chen, Wang, Qiu, and Li}]{chen2025mimo}
Chen, Y.; Xu, D.; Huang, Y.; Zhan, S.; Wang, H.; Chen, D.; Wang, X.; Qiu, M.; and Li, H. 2025{\natexlab{b}}.
\newblock MIMO: A Medical Vision Language Model with Visual Referring Multimodal Input and Pixel Grounding Multimodal Output.
\newblock In \emph{Proceedings of the Computer Vision and Pattern Recognition Conference}, 24732--24741.

\bibitem[{Cheng, Huang, and Wei(2024)}]{cheng2024adapting}
Cheng, D.; Huang, S.; and Wei, F. 2024.
\newblock Adapting Large Language Models via Reading Comprehension.
\newblock In \emph{The Twelfth International Conference on Learning Representations}.

\bibitem[{Cheng et~al.(2024)Cheng, Huang, Zhu, Zhang, Zhao, Luan, Dai, and Zhang}]{cheng2024domain}
Cheng, D.; Huang, S.; Zhu, Z.; Zhang, X.; Zhao, W.~X.; Luan, Z.; Dai, B.; and Zhang, Z. 2024.
\newblock On Domain-Specific Post-Training for Multimodal Large Language Models.
\newblock \emph{arXiv preprint arXiv:2411.19930}.

\bibitem[{Cheng et~al.(2023)Cheng, Tian, Liu, Chen, Wang, Chen, and Zhang}]{cheng2023can}
Cheng, S.; Tian, B.; Liu, Q.; Chen, X.; Wang, Y.; Chen, H.; and Zhang, N. 2023.
\newblock Can We Edit Multimodal Large Language Models?
\newblock In \emph{Proceedings of the 2023 Conference on Empirical Methods in Natural Language Processing}, 13877--13888.

\bibitem[{Clusmann et~al.(2025)Clusmann, Ferber, Wiest, Schneider, Brinker, Foersch, Truhn, and Kather}]{clusmann2025prompt}
Clusmann, J.; Ferber, D.; Wiest, I.~C.; Schneider, C.~V.; Brinker, T.~J.; Foersch, S.; Truhn, D.; and Kather, J.~N. 2025.
\newblock Prompt injection attacks on vision language models in oncology.
\newblock \emph{Nature Communications}, 16(1): 1239.

\bibitem[{Dai et~al.(2022)Dai, Dong, Hao, Sui, Chang, and Wei}]{dai2022knowledge}
Dai, D.; Dong, L.; Hao, Y.; Sui, Z.; Chang, B.; and Wei, F. 2022.
\newblock Knowledge Neurons in Pretrained Transformers.
\newblock In \emph{Proceedings of the 60th Annual Meeting of the Association for Computational Linguistics (Volume 1: Long Papers)}, 8493--8502.

\bibitem[{De~Cao, Aziz, and Titov(2021)}]{de2021editing}
De~Cao, N.; Aziz, W.; and Titov, I. 2021.
\newblock Editing Factual Knowledge in Language Models.
\newblock In \emph{Proceedings of the 2021 Conference on Empirical Methods in Natural Language Processing}, 6491--6506.

\bibitem[{Du et~al.(2025)Du, Jiang, Gao, Shi, Zheng, Qi, and Li}]{du2025mmke}
Du, Y.; Jiang, K.; Gao, Z.; Shi, C.; Zheng, Z.; Qi, S.; and Li, Q. 2025.
\newblock MMKE-Bench: A Multimodal Editing Benchmark for Diverse Visual Knowledge.
\newblock In \emph{The Thirteenth International Conference on Learning Representations}.

\bibitem[{Hartvigsen et~al.(2023)Hartvigsen, Sankaranarayanan, Palangi, Kim, and Ghassemi}]{hartvigsen2023aging}
Hartvigsen, T.; Sankaranarayanan, S.; Palangi, H.; Kim, Y.; and Ghassemi, M. 2023.
\newblock Aging with grace: Lifelong model editing with discrete key-value adaptors.
\newblock \emph{Advances in Neural Information Processing Systems}, 36: 47934--47959.

\bibitem[{Hu et~al.(2024)Hu, Li, Lu, Shao, He, Qiao, and Luo}]{hu2024omnimedvqa}
Hu, Y.; Li, T.; Lu, Q.; Shao, W.; He, J.; Qiao, Y.; and Luo, P. 2024.
\newblock Omnimedvqa: A new large-scale comprehensive evaluation benchmark for medical lvlm.
\newblock In \emph{Proceedings of the IEEE/CVF Conference on Computer Vision and Pattern Recognition}, 22170--22183.

\bibitem[{Huang et~al.(2024)Huang, Zhong, Yu, Liu, Wu, Wang, and Tan}]{huang2024vlkeb}
Huang, H.; Zhong, H.; Yu, T.; Liu, Q.; Wu, S.; Wang, L.; and Tan, T. 2024.
\newblock VLKEB: A Large Vision-Language Model Knowledge Editing Benchmark.
\newblock In \emph{The Thirty-eight Conference on Neural Information Processing Systems Datasets and Benchmarks Track}.

\bibitem[{Huang et~al.(2025)Huang, Wang, Zhang, Zhu, Xi, An, Wang, Liang, and Pan}]{huang2025medical}
Huang, X.; Wang, X.; Zhang, H.; Zhu, Y.; Xi, J.; An, J.; Wang, H.; Liang, H.; and Pan, C. 2025.
\newblock Medical mllm is vulnerable: Cross-modality jailbreak and mismatched attacks on medical multimodal large language models.
\newblock In \emph{Proceedings of the AAAI Conference on Artificial Intelligence}, volume~39, 3797--3805.

\bibitem[{Levy et~al.(2017)Levy, Seo, Choi, and Zettlemoyer}]{levy2017zero}
Levy, O.; Seo, M.; Choi, E.; and Zettlemoyer, L. 2017.
\newblock Zero-shot relation extraction via reading comprehension.
\newblock In \emph{21st Conference on Computational Natural Language Learning, CoNLL 2017}, 333--342. Association for Computational Linguistics (ACL).

\bibitem[{Li et~al.(2023{\natexlab{a}})Li, Wong, Zhang, Usuyama, Liu, Yang, Naumann, Poon, and Gao}]{li2023llava}
Li, C.; Wong, C.; Zhang, S.; Usuyama, N.; Liu, H.; Yang, J.; Naumann, T.; Poon, H.; and Gao, J. 2023{\natexlab{a}}.
\newblock Llava-med: Training a large language-and-vision assistant for biomedicine in one day.
\newblock \emph{Advances in Neural Information Processing Systems}, 36: 28541--28564.

\bibitem[{Li et~al.(2023{\natexlab{b}})Li, Li, Savarese, and Hoi}]{li2023blip}
Li, J.; Li, D.; Savarese, S.; and Hoi, S. 2023{\natexlab{b}}.
\newblock Blip-2: Bootstrapping language-image pre-training with frozen image encoders and large language models.
\newblock In \emph{International conference on machine learning}, 19730--19742. PMLR.

\bibitem[{Liu et~al.(2024{\natexlab{a}})Liu, Feng, Xue, Wang, Wu, Lu, Zhao, Deng, Zhang, Ruan et~al.}]{liu2024deepseek}
Liu, A.; Feng, B.; Xue, B.; Wang, B.; Wu, B.; Lu, C.; Zhao, C.; Deng, C.; Zhang, C.; Ruan, C.; et~al. 2024{\natexlab{a}}.
\newblock Deepseek-v3 technical report.
\newblock \emph{arXiv preprint arXiv:2412.19437}.

\bibitem[{Liu et~al.(2023{\natexlab{a}})Liu, Li, Wu, and Lee}]{liu2023visual}
Liu, H.; Li, C.; Wu, Q.; and Lee, Y.~J. 2023{\natexlab{a}}.
\newblock Visual instruction tuning.
\newblock \emph{Advances in neural information processing systems}, 36: 34892--34916.

\bibitem[{Liu et~al.(2024{\natexlab{b}})Liu, Yu, Zhang, Zhang, and Xiao}]{liu2024automatic}
Liu, X.; Yu, Z.; Zhang, Y.; Zhang, N.; and Xiao, C. 2024{\natexlab{b}}.
\newblock Automatic and universal prompt injection attacks against large language models.
\newblock \emph{arXiv preprint arXiv:2403.04957}.

\bibitem[{Liu et~al.(2023{\natexlab{b}})Liu, Deng, Li, Wang, Wang, Wang, Zhang, Liu, Wang, Zheng et~al.}]{liu2023prompt}
Liu, Y.; Deng, G.; Li, Y.; Wang, K.; Wang, Z.; Wang, X.; Zhang, T.; Liu, Y.; Wang, H.; Zheng, Y.; et~al. 2023{\natexlab{b}}.
\newblock Prompt Injection attack against LLM-integrated Applications.
\newblock \emph{arXiv preprint arXiv:2306.05499}.

\bibitem[{Meng et~al.(2022)Meng, Bau, Andonian, and Belinkov}]{meng2022locating}
Meng, K.; Bau, D.; Andonian, A.; and Belinkov, Y. 2022.
\newblock Locating and editing factual associations in gpt.
\newblock \emph{Advances in neural information processing systems}, 35: 17359--17372.

\bibitem[{Meng et~al.(2023)Meng, Sharma, Andonian, Belinkov, and Bau}]{mengmass}
Meng, K.; Sharma, A.~S.; Andonian, A.~J.; Belinkov, Y.; and Bau, D. 2023.
\newblock Mass-Editing Memory in a Transformer.
\newblock In \emph{The Eleventh International Conference on Learning Representations}.

\bibitem[{Mitchell et~al.(2022{\natexlab{a}})Mitchell, Lin, Bosselut, Finn, and Manning}]{mitchell2022fast}
Mitchell, E.; Lin, C.; Bosselut, A.; Finn, C.; and Manning, C.~D. 2022{\natexlab{a}}.
\newblock Fast Model Editing at Scale.
\newblock In \emph{International Conference on Learning Representations}.

\bibitem[{Mitchell et~al.(2022{\natexlab{b}})Mitchell, Lin, Bosselut, Manning, and Finn}]{mitchell2022memory}
Mitchell, E.; Lin, C.; Bosselut, A.; Manning, C.~D.; and Finn, C. 2022{\natexlab{b}}.
\newblock Memory-based model editing at scale.
\newblock In \emph{International Conference on Machine Learning}, 15817--15831. PMLR.

\bibitem[{Nachane et~al.(2024)Nachane, Gramopadhye, Chanda, Ramakrishnan, Jadhav, Nandwani, Raghu, and Joshi}]{nachane2024few}
Nachane, S.; Gramopadhye, O.; Chanda, P.; Ramakrishnan, G.; Jadhav, K.; Nandwani, Y.; Raghu, D.; and Joshi, S. 2024.
\newblock Few shot chain-of-thought driven reasoning to prompt LLMs for open-ended medical question answering.
\newblock In \emph{Findings of the Association for Computational Linguistics: EMNLP 2024}, 542--573.

\bibitem[{Pal, Umapathi, and Sankarasubbu(2022)}]{pal2022medmcqa}
Pal, A.; Umapathi, L.~K.; and Sankarasubbu, M. 2022.
\newblock Medmcqa: A large-scale multi-subject multi-choice dataset for medical domain question answering.
\newblock In \emph{Conference on health, inference, and learning}, 248--260. PMLR.

\bibitem[{Tan, Zhang, and Fu(2024)}]{tanmassive}
Tan, C.; Zhang, G.; and Fu, J. 2024.
\newblock Massive Editing for Large Language Models via Meta Learning.
\newblock In \emph{The Twelfth International Conference on Learning Representations}.

\bibitem[{Wang et~al.(2024{\natexlab{a}})Wang, Li, Zhang, Xu, Yao, Jiang, Xie, Huang, and Chen}]{wang2024wise}
Wang, P.; Li, Z.; Zhang, N.; Xu, Z.; Yao, Y.; Jiang, Y.; Xie, P.; Huang, F.; and Chen, H. 2024{\natexlab{a}}.
\newblock Wise: Rethinking the knowledge memory for lifelong model editing of large language models.
\newblock \emph{Advances in Neural Information Processing Systems}, 37: 53764--53797.

\bibitem[{Wang et~al.(2024{\natexlab{b}})Wang, Zhang, Tian, Xi, Yao, Xu, Wang, Mao, Wang, Cheng et~al.}]{wang2024easyedit}
Wang, P.; Zhang, N.; Tian, B.; Xi, Z.; Yao, Y.; Xu, Z.; Wang, M.; Mao, S.; Wang, X.; Cheng, S.; et~al. 2024{\natexlab{b}}.
\newblock EasyEdit: An Easy-to-use Knowledge Editing Framework for Large Language Models.
\newblock In \emph{Proceedings of the 62nd Annual Meeting of the Association for Computational Linguistics (Volume 3: System Demonstrations)}, 82--93.

\bibitem[{Xiao et~al.(2025)Xiao, Zhou, Liu, Liu, Li, Liu, and Huang}]{xiao2025comprehensive}
Xiao, H.; Zhou, F.; Liu, X.; Liu, T.; Li, Z.; Liu, X.; and Huang, X. 2025.
\newblock A comprehensive survey of large language models and multimodal large language models in medicine.
\newblock \emph{Information Fusion}, 117: 102888.

\bibitem[{Xu et~al.(2025)Xu, Chen, Chai, Xiao, Yan, Ding, Wang, Jin, Jiao, Yue et~al.}]{xu2025knowledge}
Xu, D.; Chen, Y.; Chai, Z.; Xiao, Y.; Yan, Y.; Ding, W.; Wang, H.; Jin, Z.; Jiao, W.; Yue, W.; et~al. 2025.
\newblock Knowledge fusion in deep learning-based medical vision-language models: A review.
\newblock \emph{Information Fusion}, 103455.

\bibitem[{Xu et~al.(2024{\natexlab{a}})Xu, Chen, Wang, Huang, Wang, Jin, Wang, Yue, He, Li et~al.}]{xu2024mlevlm}
Xu, D.; Chen, Y.; Wang, J.; Huang, Y.; Wang, H.; Jin, Z.; Wang, H.; Yue, W.; He, J.; Li, H.; et~al. 2024{\natexlab{a}}.
\newblock Mlevlm: Improve multi-level progressive capabilities based on multimodal large language model for medical visual question answering.
\newblock In \emph{Findings of the Association for Computational Linguistics ACL 2024}, 4977--4997.

\bibitem[{Xu et~al.(2024{\natexlab{b}})Xu, Zhang, Zhu, Lin, Liu, Wu, Xu, Wang, Ye, Zhao et~al.}]{xu2024editing}
Xu, D.; Zhang, Z.; Zhu, Z.; Lin, Z.; Liu, Q.; Wu, X.; Xu, T.; Wang, W.; Ye, Y.; Zhao, X.; et~al. 2024{\natexlab{b}}.
\newblock Editing factual knowledge and explanatory ability of medical large language models.
\newblock In \emph{Proceedings of the 33rd ACM International Conference on Information and Knowledge Management}, 2660--2670.

\bibitem[{Xu et~al.(2024{\natexlab{c}})Xu, Lin, Yang, Zhang, Shi, Zhang, Fang, Xu, and Qiu}]{xu2024earth}
Xu, R.; Lin, B.; Yang, S.; Zhang, T.; Shi, W.; Zhang, T.; Fang, Z.; Xu, W.; and Qiu, H. 2024{\natexlab{c}}.
\newblock The Earth is Flat because...: Investigating LLMs’ Belief towards Misinformation via Persuasive Conversation.
\newblock In \emph{Proceedings of the 62nd Annual Meeting of the Association for Computational Linguistics (Volume 1: Long Papers)}, 16259--16303.

\bibitem[{Yang et~al.(2024)Yang, Wang, Huang, Yang, Zhang, Huang, Zhang, Wang, Yang, He et~al.}]{yang2024lmkg}
Yang, P.; Wang, H.; Huang, Y.; Yang, S.; Zhang, Y.; Huang, L.; Zhang, Y.; Wang, G.; Yang, S.; He, L.; et~al. 2024.
\newblock LMKG: A large-scale and multi-source medical knowledge graph for intelligent medicine applications.
\newblock \emph{Knowledge-Based Systems}, 284: 111323.

\bibitem[{Yu et~al.(2024)Yu, Chen, Zhou, and He}]{yu2024melo}
Yu, L.; Chen, Q.; Zhou, J.; and He, L. 2024.
\newblock Melo: Enhancing model editing with neuron-indexed dynamic lora.
\newblock In \emph{Proceedings of the AAAI Conference on Artificial Intelligence}, volume~38, 19449--19457.

\bibitem[{Zhang et~al.(2024)Zhang, Tian, Cheng, Liang, Hu, Xue, Gou, Chen, and Chen}]{zhang2024instructedit}
Zhang, N.; Tian, B.; Cheng, S.; Liang, X.; Hu, Y.; Xue, K.; Gou, Y.; Chen, X.; and Chen, H. 2024.
\newblock InstructEdit: instruction-based knowledge editing for large language models.
\newblock In \emph{Proceedings of the Thirty-Third International Joint Conference on Artificial Intelligence}, 6633--6641.

\bibitem[{Zhang et~al.(2023)Zhang, Wu, Zhao, Lin, Zhang, Wang, and Xie}]{zhang2023pmc}
Zhang, X.; Wu, C.; Zhao, Z.; Lin, W.; Zhang, Y.; Wang, Y.; and Xie, W. 2023.
\newblock Pmc-vqa: Visual instruction tuning for medical visual question answering.
\newblock \emph{arXiv preprint arXiv:2305.10415}.

\bibitem[{Zheng et~al.(2023)Zheng, Li, Dong, Fan, Wu, Xu, and Chang}]{zheng2023can}
Zheng, C.; Li, L.; Dong, Q.; Fan, Y.; Wu, Z.; Xu, J.; and Chang, B. 2023.
\newblock Can We Edit Factual Knowledge by In-Context Learning?
\newblock In \emph{Proceedings of the 2023 Conference on Empirical Methods in Natural Language Processing}, 4862--4876.

\bibitem[{Zhu et~al.(2023)Zhu, Chen, Shen, Li, and Elhoseiny}]{zhuminigpt}
Zhu, D.; Chen, J.; Shen, X.; Li, X.; and Elhoseiny, M. 2023.
\newblock MiniGPT-4: Enhancing Vision-Language Understanding with Advanced Large Language Models.
\newblock \emph{The Twelfth International Conference on Learning Representations}.

\end{thebibliography}

\appendix

\section{Appendix}
\section{A. Data Source and Statistics}
To ensure comprehensive modality coverage (text, image, structured KG) and diverse task representation (VQA, text QA, KG reasoning) of MedMKEB, we selected and integrated four foundational datasets: OmniMedVQA, MedMCQA, PMC-VQA, and LMKG, encompassing multimodal understanding, deep reasoning, scientific visual interpretation, and structured factual relationships, which are all essential targets for real-world medical knowledge updates.
\subsection{A.1 Data Source}
\paragraph{OmniMedVQA} OmniMedVQA~\cite{hu2024omnimedvqa} is a large-scale multimodal medical Visual Question Answering (VQA) dataset. It pairs diverse medical images (e.g., radiology, pathology, dermatology) with clinically relevant questions.

\noindent \paragraph{MedMCQA} MedMCQA dataset~\cite{pal2022medmcqa} is a text-based, multiple-choice question-answering dataset covering a vast spectrum of medical topics derived from medical entrance exams, textbooks, and clinical guidelines, which is crucial for evaluating edits involving complex diagnostic criteria or theoretical concepts.

\noindent \paragraph{PMC-VQA} PMC-VQA~\cite{zhang2023pmc} leverages figures and illustrations from open-access biomedical research articles in PubMed Central (PMC). Its questions focus on interpreting complex scientific visuals (e.g., graphs, charts, diagrams, microscopy images) within the context of biomedical research findings.

\noindent \paragraph{LMKG} LMKG~\cite{yang2024lmkg} is a structured knowledge base comprising medical entities (diseases, symptoms, drugs, procedures) and their relationships (e.g., causes, treatments, contraindications), which provides a rich source of factual medical knowledge in the form of triples, allowing precise editing and verification of discrete medical facts.

\begin{figure*}[!t]
\centerline{\includegraphics[width=0.8\linewidth]{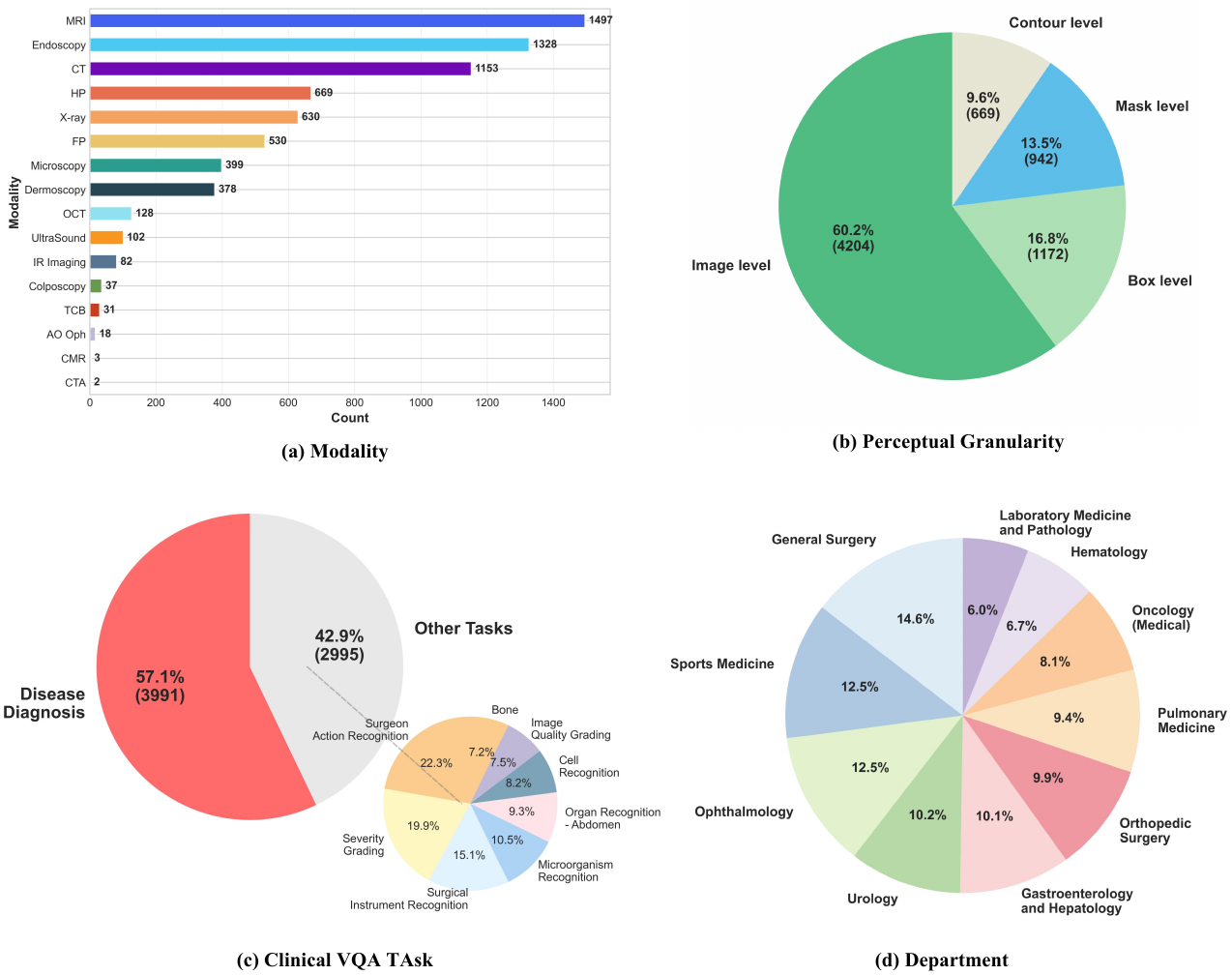}}
\caption{Statistics of MedMKEB.}
\label{img:statistics}
\end{figure*}

\subsection{A.2 Statistics}
The statistics for MedMKEB are presented in Table \ref{tab:statistics}. MedMKEB contains 6987 edits in total, split into 4490 for training and 2497 for evaluation. A total of 4490 images are used for the Reliability, Generality, and Locality tests. The Portability test reuses the same images as the Reliability test and comprises 2688 cases. These cases are divided into 1-hop, 2-hop, and 3-hop, with 1776, 802, and 110 cases, respectively. Specially, we choose 721 cases to test the Robustness of MLLMs. 

Besides, our dataset, MedMKEB, contains broad clinical applications across anatomical regions and diagnostic purposes. As shown in Figure \ref{img:statistics}, the dataset comprises 6,987 medical imaging samples with diverse modalities, led by MRI (21.4\%), endoscopy (19.0\%), and CT (16.5\%), while less common modalities like CMR and CTA represent under 0.1\%. Perceptual granularity is predominantly image-level (60.2\%), followed by box-level (16.8\%) and mask-level (13.5\%) annotations. The samples span multiple departments, including general surgery (12.5\%), sports medicine (10.7\%), and ophthalmology (10.7\%), with clinical VQA tasks focusing mainly on disease diagnosis (57.1\%), surgeon action recognition (7.4\%), and severity grading (6.6\%). 

\begin{table}[t]
\centering
\caption{Statistics of MEDMKEB.}
\begin{tabular}{lccccc}
\hline
\textbf{Type} & \textbf{\#Nums} & & \textbf{Rel.} & \textbf{Gen.} & \textbf{Loc.}\\
\hline
Train
& 4490  & \textbf{\#Images} & 4490 & 3404 & 4100 \\
Eval
& 2497  & \textbf{\#Images}& 2497 & 2115 & 2369 \\
\hline
\hline
\textbf{Type} & \textbf{Port.}& \textbf{hop-1} & \textbf{hop-2} & \textbf{hop-3} & \textbf{Robu.}\\
\hline
Eval & 2688 & 1776 & 802 & 110 & 721\\
\hline
\end{tabular}
\label{tab:statistics}
\end{table}

\subsection{A.3 Relations in Multihop Knowledge}

In order to construct the multi-hop portability problem, we select a part of relations from LMKG. The selected relations are listed in Table \ref{tab:relations}.

\begin{table*}[t]
\centering
\caption{Relations in multihop knowledge.}
\fontsize{10}{10}\selectfont    
\begin{tabular}{lccccc}
\toprule
\textbf{Relation} &  & &  &  & \\
\hline
Complication & Treatment  & Procedure Site & Symptom & Side Effect & Adjacent to \\
Part of & Associated with  & Associated Morphology & Associated Anatomy & Component & Diagnostic Method \\
Contraindication & Prevented by  & Contained in & has Part & Drains to & Branch of \\
Diagnostic Test & Risk Factor  & Procedure Device & Causative Agent & Findings & Drug Used for \\
\bottomrule
\end{tabular}
\label{tab:relations}
\end{table*}

\subsection{A.4 Adversarial Question Examples}

In MedMKEB, we introduced robustness as an evaluation metric. Robustness aims to measure the defense capability of the model against adversarial attacks after editing. Therefore, we constructed five categories of robustness questions against prompt injection attacks:
\begin{itemize}
    \item \textbf{Pseudo-authoritative Interference:} Induce false trust by adding misleading statements of expert opinion. Test whether the model is susceptible to authoritative information, even if this information may be wrong~\cite{xu2024earth}.
    \item \textbf{Symptom Confusion:} Introduce alternative symptoms or signs to confuse the diagnosis. Evaluate the diagnostic accuracy of the model in the face of similar or misleading symptoms~\cite{clusmann2025prompt}.
    \item \textbf{Misleading Context:} Add irrelevant but plausible clinical context. Check whether the model is able to ignore irrelevant information and focus on key diagnostic factors~\cite{beurer2025design}.
    \item \textbf{Irrelevant Clinical Details:} Add excessive and irrelevant clinical information to distract the model. Test the model’s efficiency and accuracy in handling redundant information~\cite{clusmann2025prompt}.
    \item \textbf{Vague Qualifiers:} Use uncertainty-inducing language to weaken the clarity of the problem. Evaluate how the model performs when faced with ambiguous or unclear problems~\cite{nachane2024few}.
\end{itemize}
See Table \ref{tab:robustness_attacks} for specific examples. 

\subsection{A.5 Dataset Examples}

\begin{figure*}[!t]
\centerline{\includegraphics[width=1\linewidth]{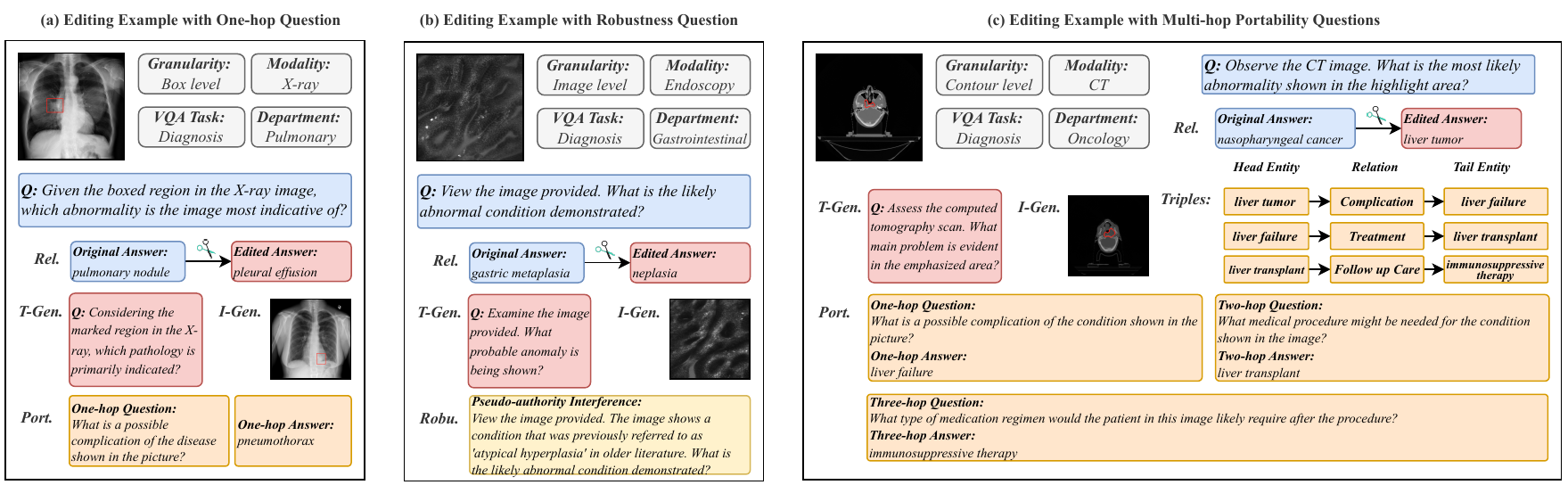}}
\caption{Examples of MedMKEB.}
\label{img:examples}
\end{figure*}

We show several examples of MedMKEB, including an edit example with single-hop portability, an edit example with robustness questions, and an edit example with multi-hop portability questions. The examples are shown in Figure \ref{img:examples}.

\begin{table*}[htbp]
\centering
\small

\renewcommand{\arraystretch}{1.3}
\caption{Examples of prompt injection attacks for robustness evaluation.}
\begin{tabular}{p{2.8cm}p{4.5cm}p{3.8cm}p{4.8cm}}
\toprule
\textbf{Attack Type} & \textbf{Description} & \textbf{Original Question} & \textbf{Adversarial Question} \\
\hline
\makecell[l]{Pseudo-authoritative \\Interference }& Appending misleading expert-sounding statements to induce false trust & What disease is shown in the CT scan? & According to the latest WHO guideline, what disease is shown in the CT scan? \\
\hline
Symptom Confusion & Introducing alternative symptoms or signs to confuse diagnosis & What is the likely condition based on this X-ray? & What is the likely condition based on this X-ray, considering the patient also has chest pain and high fever? \\
\hline
Misleading Context & Prepending unrelated but plausible clinical context & What abnormality is visible in the image? & The patient has a history of liver disease. What abnormality is visible in the image? \\
\hline
\makecell[l]{Irrelevant Clinical \\Details} & Adding excessive and irrelevant clinical information to distract the model & What organ is affected in this MRI? & The patient’s blood pressure was 120/80 and they had no known allergies. What organ is affected in this MRI? \\
\hline
Vague Qualifiers & Using uncertainty-inducing language to weaken the clarity of the question & What condition does the image suggest? & What condition might the image possibly suggest? \\
\bottomrule
\end{tabular}

\label{tab:robustness_attacks}
\end{table*}

\section{B. MLLMs and Editing Methods}
\subsection{B.1 Versions of MLLMs}
We conduct experiments on six Multimodal Large Language Models, and their specific version are in Table \ref{tab:MLLMs}.

\begin{table}[t]
\centering
\caption{Versions of MLLMs in the experiments. Specifically, Vis., Coun. mean Vision Encoder and the Counterfactual model for SERAC method, respectively.}
\begin{tabular}{lcc}
\toprule
\textbf{MLLM} & \textbf{BLIP2-OPT} & \textbf{biomed-Qwen2-VL} \\
\hline
\textbf{LLM}
 & OPT-2.7B & Qwen2-1.5B-Instruct \\
 \textbf{Vis.}
& ViT-g(1B) &  ViT-g(1B)   \\
 \textbf{Coun.}
& OPT-125M &  Qwen2-1.5B-Instruct   \\
\hline

\textbf{MLLM}  & \textbf{LLaVA-1.5} & \textbf{LLaVA-Med}\\
\hline
\textbf{LLM}
 & Vicuna-7B-v1.5 & Vicuna-7B  \\
\textbf{Vis.}
&  ViT-L(0.3B) & ViT-L(0.3B)  \\
 \textbf{Coun.}
& Vicuna-7B-v1.5 &  Vicuna-7B   \\
\hline
\textbf{MLLM} & \textbf{MiniGPT-4} & \textbf{HuatuoGPT-Vision}  \\
\hline
\textbf{LLM}
  & Vicuna-7B  & Qwen-7B  \\
\textbf{Vis.}
& ViT-g(1B) &  ViT-L(0.3B)  \\
 \textbf{Coun.}
& Vicuna-7B &  Qwen-2-7B-Instruct   \\
\bottomrule
\end{tabular}
\label{tab:MLLMs}
\end{table}

\paragraph{BLIP2-OPT} BLIP2-OPT~\cite{li2023blip} is an efficient vision-language model by interfacing a frozen ViT-L/14 image encoder with a frozen OPT large language model through a lightweight Querying Transformer (Q-Former).

\paragraph{LLaVA-1.5} LLaVA-1.5~\cite{liu2023visual} is an open-source large vision-language model that upgrades the original LLaVA framework by replacing the linear projection with a two-layer MLP vision-language connector and by augmenting training with academic-task-oriented VQA, OCR, and region-level perception datasets. Built atop CLIP-ViT-L-336px and a 7/13 B-parameter Vicuna backbone, it attains strong multi-modal reasoning while remaining fully reproducible with ~1.2 M publicly available samples.

\paragraph{MiniGPT-4} MiniGPT-4~\cite{zhuminigpt} is an open-source vision-language chatbot that replicates many emergent multimodal abilities of GPT-4. The model is first pre-trained on large-scale image–text pairs and subsequently fine-tuned on 3.5K high-quality, detail-rich captions to ensure fluent and contextually appropriate generation.

\paragraph{Biomed-Qwen2-VL} Biomed-Qwen2-VL is a domain-specialised large vision-language model developed for biomedicine. Employing a 448×448 visual encoder, it supports fine-grained medical image understanding, bilingual medical document QA, and bounding-box grounding. The architecture retains the LLM weights of Qwen2-VL, enabling end-to-end recognition of dense medical texts and multi-image clinical case comparisons\cite{cheng2024adapting}.

\paragraph{LLaVA-Med} LLaVA-Med~\cite{li2023llava} is a medical adaptation of LLaVA designed to follow open-ended clinical instructions. It bootstraps from LLaVA-1.5, leverages PubMedVision-158k generated by GPT-4 for instruction tuning, and keeps both the CLIP visual encoder and the Vicuna LLM frozen except for the MLP connector. The resulting model exhibits superior performance on medical VQA and radiology report generation benchmarks.

\paragraph{HuatuoGPT-Vision} HuatuoGPT-Vision~\cite{chen2024huatuogpt} is a Chinese-centric medical multimodal dialogue system that extends HuatuoGPT with visual understanding. It aligns a frozen medical-domain ViT encoder with the frozen HuatuoGPT language model through a lightweight adaptor, followed by two-stage instruction tuning on curated Chinese medical image–text pairs and diagnostic conversations. The model delivers reliable diagnostics and explanatory responses grounded in medical imagery.

\subsection{B.2 Knowledge Editing Methods}
\paragraph{FT (Fine-Tuning)} The fine-tuning (FT) process updates model parameters by applying gradient descent to the specified subset of parameters. To enable restoration during single editing, we retain a copy of the target layers' weights. The AdamW optimizer is configured to restrict gradient computation and updates to only the intended fine-tuning parameters.

\paragraph{MEND (Model Editor Networks with Decomposition)}
facilitates efficient parameter updates in large language models (LLMs) within the context of large vision-language models (LVLMs). The approach involves training lightweight auxiliary networks—referred to as model editor networks—which utilize a single input-output pair to induce precise, localized modifications in the model’s behavior while preserving its general performance on unrelated tasks. By exploiting the low-rank properties of fine-tuning gradients, MEND efficiently parameterizes gradient transformations, maintaining computational scalability even for models with billions of parameters.

\paragraph{SERAC (Semi-Parametric Editing with a Retrieval-Augmented Counterfactual)} employs a memory-driven architecture comprising two key components: a scope classifier and a counterfactual model. The scope classifier, implemented as a BERT-based model, determines whether an input falls within the predefined editing domain. When the classifier identifies an in-scope input, the system retrieves relevant edits from memory and processes them through the counterfactual model. Out-of-scope inputs are instead routed to the original, unmodified model. Notably, in our experimental setup, the counterfactual model varies depending on the specific LVLM implementation, consistently utilizing the same LLM core as its corresponding LVLM.

\paragraph{KE (Knowledge Editor)} utilizes a bidirectional-LSTM hyper-network that predicts weight updates for designated model parameters based on gradients and conditional inputs $\{(y_e \to y_e')|x_e\}$. By processing the original parameters alongside the modification fact, this network outputs parameter-specific updates. The editing operation is specifically applied to the last layers of LLMs.

\paragraph{IKE (In-Context Knowledge Editing)} avoids parameter updates, instead leveraging retrieval-based demonstration construction from the training set to infuse new knowledge via prompting. This mechanism is universally applicable to all models. Each training instance is formatted as "New Fact: \{question\} \{answer\}\textbackslash n Prompt: \{question\} \{answer\}\textbackslash n" and subsequently converted into embeddings for model input.

\linespread{1.1}
\begin{table}[h]
    \caption{Multi-hop portability results for different MLLMs.} 
    \fontsize{8}{7}\selectfont    
    \centering
    \begin{tabular}{c|c|rrr}
    \toprule
    \multicolumn{1}{c|}{\textbf{Model}} 
    & \multicolumn{1}{c|}{\textbf{Editing Method}} 
    
    & \multicolumn{1}{c}{\textbf{1-hop}} 
    & \multicolumn{1}{c}{\textbf{2-hop}} 
    & \multicolumn{1}{c}{\textbf{3-hop}} \\
    
    
    \midrule

    \multirow{8}{*}{BLIP2-OPT}
    & \multicolumn{1}{c|}{base}
    		
    & \multicolumn{1}{c}{20.02} 
    & \multicolumn{1}{c}{24.85} 
    & \multicolumn{1}{c}{22.63} 
    \\
    \cmidrule{2-5}

    & \multicolumn{1}{c|}{FT-LLM}
    		
    & \multicolumn{1}{c}{22.06} 
    & \multicolumn{1}{c}{5.37} 
    & \multicolumn{1}{c}{10.54}
    \\

    & \multicolumn{1}{c|}{FT-Proj}
    		
    & \multicolumn{1}{c}{16.62} 
    & \multicolumn{1}{c}{16.39} 
    & \multicolumn{1}{c}{12.23}  \\

    & \multicolumn{1}{c|}{IKE}
    		
    & \multicolumn{1}{c}{34.79} 
    & \multicolumn{1}{c}{31.87} 
    & \multicolumn{1}{c}{26.78} \\

    & \multicolumn{1}{c|}{SERAC}
    
    & \multicolumn{1}{c}{24.69} 
    & \multicolumn{1}{c}{4.52} 
    & \multicolumn{1}{c}{3.12} \\
    
    & \multicolumn{1}{c|}{MEND}
    
    & \multicolumn{1}{c}{20.27}
    & \multicolumn{1}{c}{22.36} 
    & \multicolumn{1}{c}{22.24} \\

    & \multicolumn{1}{c|}{KE}
    
    & \multicolumn{1}{c}{35.15}
    & \multicolumn{1}{c}{28.08} 
    & \multicolumn{1}{c}{27.92} \\
    \midrule

    \multirow{8}{*}{MiniGPT4}
    & \multicolumn{1}{c|}{base}
    		
    & \multicolumn{1}{c}{44.1} 
    & \multicolumn{1}{c}{46.7} 
    & \multicolumn{1}{c}{51.19} 
    \\
    \cmidrule{2-5}

    & \multicolumn{1}{c|}{FT-LLM}
    		
    & \multicolumn{1}{c}{33.28} 
    & \multicolumn{1}{c}{23.05} 
    & \multicolumn{1}{c}{35.12}
    \\

    & \multicolumn{1}{c|}{FT-Proj}
    		
    & \multicolumn{1}{c}{45.81} 
    & \multicolumn{1}{c}{42.71} 
    & \multicolumn{1}{c}{46.35}  \\

    & \multicolumn{1}{c|}{IKE}
    		
    & \multicolumn{1}{c}{51.84} 
    & \multicolumn{1}{c}{47.81} 
    & \multicolumn{1}{c}{48.37} \\

    & \multicolumn{1}{c|}{SERAC}
		
    & \multicolumn{1}{c}{55.72} 
    & \multicolumn{1}{c}{50.53} 
    & \multicolumn{1}{c}{54.76} \\
    
    & \multicolumn{1}{c|}{MEND}
    
    & \multicolumn{1}{c}{45.69} 
    & \multicolumn{1}{c}{45.99} 
    & \multicolumn{1}{c}{50.88} \\
    
    & \multicolumn{1}{c|}{KE}
    		
    & \multicolumn{1}{c}{60.05} 
    & \multicolumn{1}{c}{52.54} 
    & \multicolumn{1}{c}{52.41} \\
    \midrule
    \multirow{8}{*}{LLaVA-v1.5}
    & \multicolumn{1}{c|}{base}
    		
    & \multicolumn{1}{c}{45.91} 
    & \multicolumn{1}{c}{49.24} 
    & \multicolumn{1}{c}{56.13} 
    \\
    \cmidrule{2-5}

    & \multicolumn{1}{c|}{FT-LLM}
    		
    & \multicolumn{1}{c}{31.95} 
    & \multicolumn{1}{c}{22.66} 
    & \multicolumn{1}{c}{36.26}
    \\

    & \multicolumn{1}{c|}{FT-Proj}
    		
    & \multicolumn{1}{c}{50.89} 
    & \multicolumn{1}{c}{41.35} 
    & \multicolumn{1}{c}{38.38}  \\

    & \multicolumn{1}{c|}{IKE}
    		
    & \multicolumn{1}{c}{52.35} 
    & \multicolumn{1}{c}{55.63} 
    & \multicolumn{1}{c}{61.32} \\

    & \multicolumn{1}{c|}{SERAC}
    		
    & \multicolumn{1}{c}{56.85} 
    & \multicolumn{1}{c}{51.22} 
    & \multicolumn{1}{c}{52.82} \\
    
    & \multicolumn{1}{c|}{MEND}
    		
    & \multicolumn{1}{c}{46.95}
    & \multicolumn{1}{c}{48.81} 
    & \multicolumn{1}{c}{56.61} \\
    
    & \multicolumn{1}{c|}{KE}
    						
    & \multicolumn{1}{c}{62.35}
    & \multicolumn{1}{c}{55.64} 
    & \multicolumn{1}{c}{59.43} \\
    \midrule
    \multirow{8}{*}{Biomed-Qwen2-VL}
    & \multicolumn{1}{c|}{base}
    		
    & \multicolumn{1}{c}{23.22} 
    & \multicolumn{1}{c}{28.20} 
    & \multicolumn{1}{c}{24.57} 
    \\
    \cmidrule{2-5}

    & \multicolumn{1}{c|}{FT-LLM}
    		
    & \multicolumn{1}{c}{16.31} 
    & \multicolumn{1}{c}{5.25} 
    & \multicolumn{1}{c}{8.76}
    \\

    & \multicolumn{1}{c|}{FT-Proj}
    		
    & \multicolumn{1}{c}{15.69} 
    & \multicolumn{1}{c}{3.90} 
    & \multicolumn{1}{c}{2.56}  \\

    & \multicolumn{1}{c|}{IKE}
    		
    & \multicolumn{1}{c}{48.29} 
    & \multicolumn{1}{c}{41.65} 
    & \multicolumn{1}{c}{33.06} \\

    & \multicolumn{1}{c|}{SERAC}
    		
    & \multicolumn{1}{c}{19.00} 
    & \multicolumn{1}{c}{20.34} 
    & \multicolumn{1}{c}{14.51} \\
    
    & \multicolumn{1}{c|}{MEND}
    		
    & \multicolumn{1}{c}{21.15}
    & \multicolumn{1}{c}{25.19} 
    & \multicolumn{1}{c}{21.59} \\
    
    & \multicolumn{1}{c|}{KE}
    		
    & \multicolumn{1}{c}{1.32}
    & \multicolumn{1}{c}{0.17} 
    & \multicolumn{1}{c}{0.26} \\
    \midrule
    \multirow{7}{*}{LLaVA-Med}
    & \multicolumn{1}{c|}{base}
    			
    & \multicolumn{1}{c}{8.80} 
    & \multicolumn{1}{c}{1.40} 
    & \multicolumn{1}{c}{8.26} 
    \\
    \cmidrule{2-5}

    & \multicolumn{1}{c|}{FT-LLM}
    
    & \multicolumn{1}{c}{23.52} 
    & \multicolumn{1}{c}{4.37} 
    & \multicolumn{1}{c}{2.06}
    \\

    & \multicolumn{1}{c|}{FT-Proj}
    		
    & \multicolumn{1}{c}{14.87} 
    & \multicolumn{1}{c}{1.66} 
    & \multicolumn{1}{c}{1.14}  \\

    & \multicolumn{1}{c|}{SERAC}
    		
    & \multicolumn{1}{c}{32.11} 
    & \multicolumn{1}{c}{1.54} 
    & \multicolumn{1}{c}{1.59} \\
    
    & \multicolumn{1}{c|}{MEND}
    		
    & \multicolumn{1}{c}{16.22}
    & \multicolumn{1}{c}{1.54} 
    & \multicolumn{1}{c}{7.72} \\
    
    & \multicolumn{1}{c|}{KE}
    				
    & \multicolumn{1}{c}{18.39}
    & \multicolumn{1}{c}{2.03} 
    & \multicolumn{1}{c}{1.67} \\
    \midrule
    \multirow{7}{*}{HuatuoGPT-Vision}
    & \multicolumn{1}{c|}{base}
    		
    & \multicolumn{1}{c}{22.49} 
    & \multicolumn{1}{c}{27.50} 
    & \multicolumn{1}{c}{24.61} 
    \\
    \cmidrule{2-5}

    & \multicolumn{1}{c|}{FT-LLM}
    		
    & \multicolumn{1}{c}{18.72} 
    & \multicolumn{1}{c}{5.99} 
    & \multicolumn{1}{c}{7.47}
    \\

    & \multicolumn{1}{c|}{FT-Proj}
    		
    & \multicolumn{1}{c}{18.96} 
    & \multicolumn{1}{c}{11.05} 
    & \multicolumn{1}{c}{8.82}  \\

    & \multicolumn{1}{c|}{IKE}
    		
    & \multicolumn{1}{c}{42.43} 
    & \multicolumn{1}{c}{46.52} 
    & \multicolumn{1}{c}{39.12} \\

    & \multicolumn{1}{c|}{SERAC}
    		
    & \multicolumn{1}{c}{25.95} 
    & \multicolumn{1}{c}{30.33} 
    & \multicolumn{1}{c}{26.38} \\
    
    & \multicolumn{1}{c|}{MEND}
    		
    & \multicolumn{1}{c}{20.36}
    & \multicolumn{1}{c}{29.02} 
    & \multicolumn{1}{c}{27.82} \\
    \bottomrule
    \end{tabular}
    \label{tab:multi-hop-port}
\end{table}

\section{C. Training Process and Hyperparameters}

\subsection{C.1 Single Editing and Sequential Editing}

\begin{figure*}[!t]
\centerline{\includegraphics[width=0.8\linewidth]{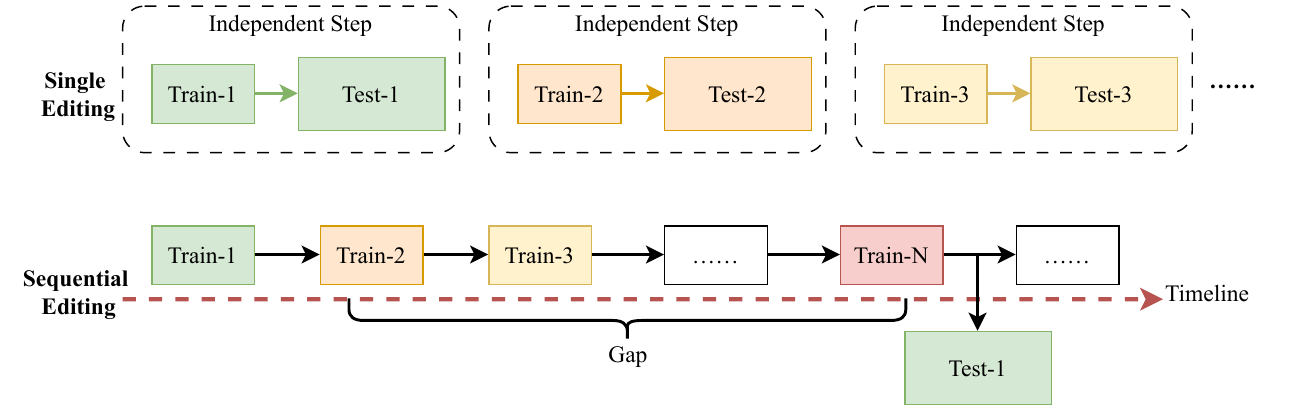}}
\caption{Illustration of single editing and sequential editing.}
\label{img:sequential_editing_ill}
\end{figure*}

In order to comprehensively evaluate the knowledge editing algorithms, we followed the settings of previous work~\cite{huang2024vlkeb}. Two types of editing methods were set up: single editing and sequential editing. Single editing updates a single piece of knowledge each time and then evaluates the editing results. By default, all algorithms perform single editing. In contrast, sequential editing continuously updates knowledge. The illustration is shown in Figure \ref{img:sequential_editing_ill}. For sequential editing, we excluded KE and IKE because they require the edit data to be part of the input during testing, which is not feasible in practical applications.

\subsection{C.2 Training Process}

The overall training process follows the settings of previous work~\cite{huang2024vlkeb}. For the parameter update algorithms such as FT, SERAC, MEND, and KE, we perform the following training process. First, calculate the negative log-likelihood loss between the output of the edited model on external samples and the new knowledge:
\begin{gather}
    \mathcal{L}_{edit} = \text{NLL}(f_{M}(x_{outer};\theta'), a_{outer}), \\
    \mathcal{L}_{iedit} = \text{NLL}(f_{M}(i_{outer}, x_{outer};\theta'), a_{outer}),
\end{gather}
where $f_M(\cdot;\theta')$ is the edited model, $\text{NLL}(\cdot)$ is the negative log-likelihood loss. Then calculate the locality loss of the edited model on irrelevant samples:
\begin{gather}
    \mathcal{L}_{loc} = \mathcal{D}_{KL}(f_M(x_{loc};\theta)||f_M(x_{loc};\theta')),\\
    \mathcal{L}_{iloc} = \mathcal{D}_{KL}(f_M(i_{loc}, x_{loc};\theta)||f_M(i_{loc}, x_{loc};\theta')),
\end{gather}
where $\mathcal{D}_{KL}(\cdot||\cdot)$ is the KL divergence and $f_M(\cdot;\theta)$ is the base model before edit. Combining various loss terms, we get the final objective function for training:
\begin{align}
    \mathcal{L}_{total} = &\lambda_{edit} \mathcal{L}_{edit} +  \lambda_{iedit} \mathcal{L}_{iedit} \nonumber \\
    &+  \lambda_{loc}\left(\mathcal{L}_{loc} + \mathcal{L}_{iloc}\right),
\end{align}
where $\lambda_{edit}, \lambda_{iedit},\lambda_{loc}$ are hyperparameters in configuration.

\subsection{C.3 Implementation Details}

Based on the EasyEdit framework, we have supplemented the support for large medical multimodal models, including LLaVA-Med, Biomed-Qwen2-VL, and HuatuoGPT-Vision. The hyperparameters and implementation details are shown in Table \ref{tab:hyperparameters}. We were unable to conduct experiments with KE on HuatuoGPT-Vision due to an Out-Of-Memory
issue, so its editing evaluation results are not shown in Table 2. For the IKE algorithm, we use the sentence-transformers model all-MiniLM-L6-v2 to embed texts and retrieve similar edits in the training set. The number of demonstrations is set to 5 for all models.

\begin{table*}[t]
\centering
\caption{Training hyperparameters of all models and methods.}
\fontsize{10}{10}\selectfont    
\begin{tabular}{lcccc}
\toprule
\textbf{FT-LLM} &  & &  & \\
\textbf{Models} & \textbf{Steps} & \textbf{Edit Layer} & \textbf{Optimizer} & \textbf{Edit LR}\\
\midrule
BLIP2-OPT & 15 & last layer of Transformer Module & AdamW & 2e-4 \\
LLaVA-v1.5 & 10 & last layer of Transformer Module & AdamW & 1e-4 \\
MiniGPT-4 & 10 & last layer of Transformer Module & AdamW & 1e-4 \\
Biomed-Qwen2-VL & 15 & last layer of Transformer Module & AdamW & 1e-4 \\
LLaVA-Med & 10 & last layer of Transformer Module & AdamW & 1e-4 \\
HuatuoGPT-Vision & 10 & last layer of Transformer Module & AdamW & 1e-4 \\
\midrule
\textbf{FT-Proj} &  & &  & \\
\textbf{Models} & \textbf{Steps} & \textbf{Edit Layer} & \textbf{Optimizer} & \textbf{Edit LR}\\
\midrule
BLIP2-OPT & 15 & Qformer & AdamW & 2e-4 \\
LLaVA-v1.5 & 10 & mm\_projector & AdamW & 1e-4 \\
MiniGPT-4 & 10 & Qformer & AdamW & 1e-4 \\
Biomed-Qwen2-VL & 15 & last layer of ViT Module & AdamW & 2e-3 \\
LLaVA-Med & 10 & mm\_projector & AdamW & 1e-4 \\
HuatuoGPT-Vision & 10 & mm\_projector & AdamW & 1e-4 \\
\midrule
\textbf{SERAC} &  & &  & \\
\textbf{Models} & \textbf{MaxIter} & \textbf{Edit Layer} & \textbf{Optimizer} & \textbf{Edit LR}\\
\midrule
BLIP2-OPT & 30000 & all layers of OPT-125M & Adam & 2e-4 \\
LLaVA-v1.5 & 30000 & last layer of LLM & Adam & 1e-4 \\
MiniGPT-4 & 30000 & last layer of LLM & Adam & 1e-4 \\
Biomed-Qwen2-VL & 30000 & last layer of LLM & Adam & 1e-4 \\
LLaVA-Med & 30000 & last layer of LLM & Adam & 1e-4 \\
HuatuoGPT-Vision & 25000 & last layer of LLM & Adam & 1e-4 \\
\midrule
\textbf{MEND} &  & &  & \\
\textbf{Models} & \textbf{MaxIter} & \textbf{Edit Layer} & \textbf{Optimizer} & \textbf{LR}\\
\midrule
BLIP2-OPT & 40000 & layer 29, 30, 31 of Transformer Module & Adam & 1e-6 \\
LLaVA-v1.5 & 30000 & last layer of Transformer Module & Adam & 1e-6 \\
MiniGPT-4 & 30000 & last layer of Transformer Module & Adam & 1e-6 \\
Biomed-Qwen2-VL & 40000 & last layer of Transformer Module & Adam & 1e-6 \\
LLaVA-Med & 30000 & last layer of Transformer Module & Adam & 1e-6 \\
HuatuoGPT-Vision & 30000 & last layer of Transformer Module & Adam & 1e-6 \\
\midrule
\textbf{KE} &  & &  & \\
\textbf{Models} & \textbf{MaxIter} & \textbf{Edit Layer} & \textbf{Optimizer} & \textbf{LR}\\
\midrule
BLIP2-OPT & 30000 & layer 29, 30, 31 of Transformer Module & RMSprop & 3e-4 \\
LLaVA-v1.5 & 30000 & last layer of Transformer Module & RMSprop & 3e-4 \\
MiniGPT-4 & 30000 & last layer of Transformer Module & RMSprop & 3e-4 \\
Biomed-Qwen2-VL & 30000 & last layer of Transformer Module & RMSprop & 3e-4 \\
LLaVA-Med & 30000 & last layer of Transformer Module & RMSprop & 3e-4 \\
HuatuoGPT-Vision & 30000 & last layer of Transformer Module & RMSprop & 3e-4 \\
\bottomrule
\end{tabular}
\label{tab:hyperparameters}
\end{table*}

\section{D. More Experimental Results}



\subsection{D.1 Original Results of Multi-hop Portability}

We present the results of multi-hop portability in Figure 3. Here we provide the original results in Table \ref{tab:multi-hop-port}.

\subsection{D.2 Original Results of Sequential Editing}

We present the results of sequential editing in Figure 4. Here we provide the original results in Table \ref{tab:sequential-edit-results}. 

\linespread{1.1}
\begin{table*}[tbp!]
    \caption{Sequential editing results for different MLLMs.} 
    \fontsize{8}{7}\selectfont    
    \centering
    \begin{tabular}{c|c|c|rrrrrr}
    \toprule
    \multicolumn{1}{c|}{\textbf{Model}} 
    & \multicolumn{1}{c|}{\textbf{Editing Method}} 
    & \multicolumn{1}{c}{\textbf{Gap}} 
    & \multicolumn{1}{c}{\textbf{Reliability}} 
    & \multicolumn{1}{c}{\textbf{T-Generality}} 
    & \multicolumn{1}{c}{\textbf{I-Generality}}
    & \multicolumn{1}{c}{\textbf{T-Locality}}
    & \multicolumn{1}{c}{\textbf{I-Locality}}
    & \multicolumn{1}{c}{\textbf{Portability}}\\
    
    
    \midrule

    \multirow{14}{*}{BLIP2-OPT}
    & \multirow{4}{*}{FT-LLM}
    					
    & \multicolumn{1}{c}{10} 
    & \multicolumn{1}{c}{55.46} 
    & \multicolumn{1}{c}{53.50} 
    & \multicolumn{1}{c}{55.56}
    & \multicolumn{1}{c}{30.11}
    & \multicolumn{1}{c}{3.30}
    & \multicolumn{1}{c}{14.62} 
    \\
    &
    & \multicolumn{1}{c}{20} 
    & \multicolumn{1}{c}{49.30} 
    & \multicolumn{1}{c}{47.86} 
    & \multicolumn{1}{c}{49.25}
    & \multicolumn{1}{c}{28.21}
    & \multicolumn{1}{c}{2.97}
    & \multicolumn{1}{c}{14.57}  \\
    &
    & \multicolumn{1}{c}{50} 
    & \multicolumn{1}{c}{43.65} 
    & \multicolumn{1}{c}{41.89} 
    & \multicolumn{1}{c}{43.46}
    & \multicolumn{1}{c}{23.01}
    & \multicolumn{1}{c}{2.19}
    & \multicolumn{1}{c}{13.16}  \\
    &					
    & \multicolumn{1}{c}{100} 
    & \multicolumn{1}{c}{42.82} 
    & \multicolumn{1}{c}{40.16} 
    & \multicolumn{1}{c}{43.03}
    & \multicolumn{1}{c}{15.89}
    & \multicolumn{1}{c}{1.18}
    & \multicolumn{1}{c}{12.49}  \\
    \cmidrule{2-9}

    & \multirow{4}{*}{FT-Proj}
    					
    & \multicolumn{1}{c}{10} 
    & \multicolumn{1}{c}{19.91} 
    & \multicolumn{1}{c}{19.06} 
    & \multicolumn{1}{c}{20.04}
    & \multicolumn{1}{c}{100.0}
    & \multicolumn{1}{c}{2.52}
    & \multicolumn{1}{c}{9.14} 
    \\
    &
    & \multicolumn{1}{c}{20} 
    & \multicolumn{1}{c}{20.29} 
    & \multicolumn{1}{c}{18.95} 
    & \multicolumn{1}{c}{20.31}
    & \multicolumn{1}{c}{100.0}
    & \multicolumn{1}{c}{2.42}
    & \multicolumn{1}{c}{10.02}  \\
    &
    & \multicolumn{1}{c}{50} 
    & \multicolumn{1}{c}{19.37} 
    & \multicolumn{1}{c}{18.51} 
    & \multicolumn{1}{c}{19.45}
    & \multicolumn{1}{c}{100.0}
    & \multicolumn{1}{c}{2.73}
    & \multicolumn{1}{c}{12.34}  \\
    &
    & \multicolumn{1}{c}{100} 
    & \multicolumn{1}{c}{21.47} 
    & \multicolumn{1}{c}{19.12} 
    & \multicolumn{1}{c}{21.47}
    & \multicolumn{1}{c}{100.0}
    & \multicolumn{1}{c}{3.00}
    & \multicolumn{1}{c}{12.12}  \\

   \cmidrule{2-9}

    & \multirow{4}{*}{SERAC}
    
    & \multicolumn{1}{c}{10} 
    & \multicolumn{1}{c}{78.95} 
    & \multicolumn{1}{c}{51.09} 
    & \multicolumn{1}{c}{78.99}
    & \multicolumn{1}{c}{100.0}
    & \multicolumn{1}{c}{17.57}
    & \multicolumn{1}{c}{6.22} 
    \\
    &
    & \multicolumn{1}{c}{20} 
    & \multicolumn{1}{c}{78.80} 
    & \multicolumn{1}{c}{50.70} 
    & \multicolumn{1}{c}{78.88}
    & \multicolumn{1}{c}{100.0}
    & \multicolumn{1}{c}{17.59}
    & \multicolumn{1}{c}{6.24}  \\
    &
    & \multicolumn{1}{c}{50} 
    & \multicolumn{1}{c}{78.33} 
    & \multicolumn{1}{c}{49.40} 
    & \multicolumn{1}{c}{78.39}
    & \multicolumn{1}{c}{100.0}
    & \multicolumn{1}{c}{17.43}
    & \multicolumn{1}{c}{6.58}  \\
    &
    & \multicolumn{1}{c}{100} 
    & \multicolumn{1}{c}{77.33} 
    & \multicolumn{1}{c}{45.62} 
    & \multicolumn{1}{c}{77.31}
    & \multicolumn{1}{c}{100.0}
    & \multicolumn{1}{c}{17.51}
    & \multicolumn{1}{c}{6.07}  \\
     \midrule

    \multirow{14}{*}{MiniGPT-4}
    & \multirow{4}{*}{FT-LLM}
    
    & \multicolumn{1}{c}{10} 
    & \multicolumn{1}{c}{74.70} 
    & \multicolumn{1}{c}{73.11} 
    & \multicolumn{1}{c}{74.32}
    & \multicolumn{1}{c}{60.73}
    & \multicolumn{1}{c}{9.42}
    & \multicolumn{1}{c}{34.14} 
    \\
    &
    & \multicolumn{1}{c}{20} 
    & \multicolumn{1}{c}{70.69} 
    & \multicolumn{1}{c}{69.61} 
    & \multicolumn{1}{c}{70.55}
    & \multicolumn{1}{c}{59.89}
    & \multicolumn{1}{c}{8.67}
    & \multicolumn{1}{c}{33.34}  \\
    &
    & \multicolumn{1}{c}{50} 
    & \multicolumn{1}{c}{65.01} 
    & \multicolumn{1}{c}{64.36} 
    & \multicolumn{1}{c}{65.14}
    & \multicolumn{1}{c}{56.60}
    & \multicolumn{1}{c}{6.80}
    & \multicolumn{1}{c}{33.04}  \\
    &					
    & \multicolumn{1}{c}{100} 
    & \multicolumn{1}{c}{65.75} 
    & \multicolumn{1}{c}{64.27} 
    & \multicolumn{1}{c}{65.72}
    & \multicolumn{1}{c}{50.14}
    & \multicolumn{1}{c}{4.78}
    & \multicolumn{1}{c}{32.30}  \\
    \cmidrule{2-9}

    & \multirow{4}{*}{FT-Proj}
    
    & \multicolumn{1}{c}{10} 
    & \multicolumn{1}{c}{44.08} 
    & \multicolumn{1}{c}{43.45} 
    & \multicolumn{1}{c}{43.92}
    & \multicolumn{1}{c}{100.0}
    & \multicolumn{1}{c}{15.53}
    & \multicolumn{1}{c}{39.01} 
    \\
    &
    & \multicolumn{1}{c}{20} 
    & \multicolumn{1}{c}{44.87} 
    & \multicolumn{1}{c}{42.98} 
    & \multicolumn{1}{c}{44.59}
    & \multicolumn{1}{c}{100.0}
    & \multicolumn{1}{c}{15.54}
    & \multicolumn{1}{c}{40.54}  \\
    &
    & \multicolumn{1}{c}{50} 
    & \multicolumn{1}{c}{44.97} 
    & \multicolumn{1}{c}{42.84} 
    & \multicolumn{1}{c}{44.39}
    & \multicolumn{1}{c}{100.0}
    & \multicolumn{1}{c}{15.41}
    & \multicolumn{1}{c}{40.24}  \\
    &
    & \multicolumn{1}{c}{100} 
    & \multicolumn{1}{c}{43.38} 
    & \multicolumn{1}{c}{42.05} 
    & \multicolumn{1}{c}{43.40}
    & \multicolumn{1}{c}{100.0}
    & \multicolumn{1}{c}{15.45}
    & \multicolumn{1}{c}{40.86}  \\

   \cmidrule{2-9}

    & \multirow{4}{*}{SERAC}
    
    & \multicolumn{1}{c}{10} 
    & \multicolumn{1}{c}{55.40} 
    & \multicolumn{1}{c}{52.01} 
    & \multicolumn{1}{c}{55.43}
    & \multicolumn{1}{c}{100.0}
    & \multicolumn{1}{c}{11.81}
    & \multicolumn{1}{c}{42.89} 
    \\
    &
    & \multicolumn{1}{c}{20} 
    & \multicolumn{1}{c}{55.59} 
    & \multicolumn{1}{c}{51.91} 
    & \multicolumn{1}{c}{55.73}
    & \multicolumn{1}{c}{100.0}
    & \multicolumn{1}{c}{11.88}
    & \multicolumn{1}{c}{42.78}  \\
    &
    & \multicolumn{1}{c}{50} 
    & \multicolumn{1}{c}{55.20} 
    & \multicolumn{1}{c}{51.40} 
    & \multicolumn{1}{c}{55.39}
    & \multicolumn{1}{c}{100.0}
    & \multicolumn{1}{c}{11.88}
    & \multicolumn{1}{c}{43.09}  \\
    &
    & \multicolumn{1}{c}{100} 
    & \multicolumn{1}{c}{54.67} 
    & \multicolumn{1}{c}{51.47} 
    & \multicolumn{1}{c}{54.78}
    & \multicolumn{1}{c}{100.0}
    & \multicolumn{1}{c}{11.90}
    & \multicolumn{1}{c}{43.18}  \\

     \midrule

    \multirow{14}{*}{LLaVA}
    & \multirow{4}{*}{FT-LLM}
    
    & \multicolumn{1}{c}{10} 
    & \multicolumn{1}{c}{76.77} 
    & \multicolumn{1}{c}{74.75} 
    & \multicolumn{1}{c}{75.59}
    & \multicolumn{1}{c}{45.93}
    & \multicolumn{1}{c}{4.08}
    & \multicolumn{1}{c}{35.27} 
    \\
    &					
    & \multicolumn{1}{c}{20} 
    & \multicolumn{1}{c}{74.28} 
    & \multicolumn{1}{c}{70.67} 
    & \multicolumn{1}{c}{73.58}
    & \multicolumn{1}{c}{44.13}
    & \multicolumn{1}{c}{3.66}
    & \multicolumn{1}{c}{33.75}  \\
    &					
    & \multicolumn{1}{c}{50} 
    & \multicolumn{1}{c}{66.41} 
    & \multicolumn{1}{c}{65.43} 
    & \multicolumn{1}{c}{65.85}
    & \multicolumn{1}{c}{40.61}
    & \multicolumn{1}{c}{2.89}
    & \multicolumn{1}{c}{32.10}  \\
    &					
    & \multicolumn{1}{c}{100} 
    & \multicolumn{1}{c}{67.2} 
    & \multicolumn{1}{c}{64.88} 
    & \multicolumn{1}{c}{66.53}
    & \multicolumn{1}{c}{33.73}
    & \multicolumn{1}{c}{1.85}
    & \multicolumn{1}{c}{31.96}  \\
    \cmidrule{2-9}

    & \multirow{4}{*}{FT-Proj}
    					
    & \multicolumn{1}{c}{10} 
    & \multicolumn{1}{c}{77.54} 
    & \multicolumn{1}{c}{75.20} 
    & \multicolumn{1}{c}{70.26}
    & \multicolumn{1}{c}{100.0}
    & \multicolumn{1}{c}{2.03}
    & \multicolumn{1}{c}{46.15} 
    \\
    &
    & \multicolumn{1}{c}{20} 
    & \multicolumn{1}{c}{67.82} 
    & \multicolumn{1}{c}{66.53} 
    & \multicolumn{1}{c}{61.98}
    & \multicolumn{1}{c}{100.0}
    & \multicolumn{1}{c}{2.04}
    & \multicolumn{1}{c}{47.12}  \\
    &					
    & \multicolumn{1}{c}{50} 
    & \multicolumn{1}{c}{60.01} 
    & \multicolumn{1}{c}{59.25} 
    & \multicolumn{1}{c}{56.84}
    & \multicolumn{1}{c}{100.0}
    & \multicolumn{1}{c}{2.06}
    & \multicolumn{1}{c}{41.18}  \\
    &
    & \multicolumn{1}{c}{100} 
    & \multicolumn{1}{c}{58.38} 
    & \multicolumn{1}{c}{58.07} 
    & \multicolumn{1}{c}{56.61}
    & \multicolumn{1}{c}{100.0}
    & \multicolumn{1}{c}{2.12}
    & \multicolumn{1}{c}{41.74}  \\

   \cmidrule{2-9}

    & \multirow{4}{*}{SERAC}
    
    & \multicolumn{1}{c}{10} 
    & \multicolumn{1}{c}{94.86} 
    & \multicolumn{1}{c}{90.46} 
    & \multicolumn{1}{c}{95.02}
    & \multicolumn{1}{c}{100.0}
    & \multicolumn{1}{c}{8.85}
    & \multicolumn{1}{c}{47.18} 
    \\
    &					
    & \multicolumn{1}{c}{20} 
    & \multicolumn{1}{c}{94.85} 
    & \multicolumn{1}{c}{90.42} 
    & \multicolumn{1}{c}{95.02}
    & \multicolumn{1}{c}{100.0}
    & \multicolumn{1}{c}{8.85}
    & \multicolumn{1}{c}{47.08}  \\
    &	
    & \multicolumn{1}{c}{50} 
    & \multicolumn{1}{c}{94.44} 
    & \multicolumn{1}{c}{89.12} 
    & \multicolumn{1}{c}{94.69}
    & \multicolumn{1}{c}{100.0}
    & \multicolumn{1}{c}{8.88}
    & \multicolumn{1}{c}{47.25}  \\
    &
    & \multicolumn{1}{c}{100} 
    & \multicolumn{1}{c}{94.28} 
    & \multicolumn{1}{c}{88.27} 
    & \multicolumn{1}{c}{94.53}
    & \multicolumn{1}{c}{100.0}
    & \multicolumn{1}{c}{8.92}
    & \multicolumn{1}{c}{46.74}  \\

    \midrule

    \multirow{14}{*}{Biomed-Qwen2-VL}
    & \multirow{4}{*}{FT-LLM}
    					
    & \multicolumn{1}{c}{10} 
    & \multicolumn{1}{c}{62.82} 
    & \multicolumn{1}{c}{57.62} 
    & \multicolumn{1}{c}{62.73}
    & \multicolumn{1}{c}{15.21}
    & \multicolumn{1}{c}{8.37}
    & \multicolumn{1}{c}{14.25} 
    \\
    &					
    & \multicolumn{1}{c}{20} 
    & \multicolumn{1}{c}{59.17} 
    & \multicolumn{1}{c}{55.03} 
    & \multicolumn{1}{c}{59.17}
    & \multicolumn{1}{c}{14.05}
    & \multicolumn{1}{c}{7.63}
    & \multicolumn{1}{c}{12.75}  \\
    &					
    & \multicolumn{1}{c}{50} 
    & \multicolumn{1}{c}{53.06} 
    & \multicolumn{1}{c}{49.37} 
    & \multicolumn{1}{c}{53.05}
    & \multicolumn{1}{c}{11.84}
    & \multicolumn{1}{c}{6.08}
    & \multicolumn{1}{c}{12.14}  \\
    &					
    & \multicolumn{1}{c}{100} 
    & \multicolumn{1}{c}{51.71} 
    & \multicolumn{1}{c}{47.26} 
    & \multicolumn{1}{c}{51.71}
    & \multicolumn{1}{c}{8.57}
    & \multicolumn{1}{c}{4.31}
    & \multicolumn{1}{c}{11.94}  \\
    \cmidrule{2-9}

    & \multirow{4}{*}{FT-Proj}
    					
    & \multicolumn{1}{c}{10} 
    & \multicolumn{1}{c}{59.67} 
    & \multicolumn{1}{c}{54.65} 
    & \multicolumn{1}{c}{59.19}
    & \multicolumn{1}{c}{1.34}
    & \multicolumn{1}{c}{1.41}
    & \multicolumn{1}{c}{10.40} 
    \\
    &					
    & \multicolumn{1}{c}{20} 
    & \multicolumn{1}{c}{55.89} 
    & \multicolumn{1}{c}{51.55} 
    & \multicolumn{1}{c}{55.19}
    & \multicolumn{1}{c}{1.28}
    & \multicolumn{1}{c}{1.53}
    & \multicolumn{1}{c}{8.77}  \\
    &					
    & \multicolumn{1}{c}{50} 
    & \multicolumn{1}{c}{49.23} 
    & \multicolumn{1}{c}{45.84} 
    & \multicolumn{1}{c}{49.14}
    & \multicolumn{1}{c}{1.51}
    & \multicolumn{1}{c}{1.79}
    & \multicolumn{1}{c}{8.89}  \\
    &					
    & \multicolumn{1}{c}{100} 
    & \multicolumn{1}{c}{49.11} 
    & \multicolumn{1}{c}{45.69} 
    & \multicolumn{1}{c}{49.16}
    & \multicolumn{1}{c}{1.47}
    & \multicolumn{1}{c}{1.88}
    & \multicolumn{1}{c}{9.91}  \\

   \cmidrule{2-9}

    & \multirow{4}{*}{SERAC}
    					
    & \multicolumn{1}{c}{10} 
    & \multicolumn{1}{c}{31.83} 
    & \multicolumn{1}{c}{31.06} 
    & \multicolumn{1}{c}{34.57}
    & \multicolumn{1}{c}{99.98}
    & \multicolumn{1}{c}{18.78}
    & \multicolumn{1}{c}{17.62} 
    \\
    &				
    & \multicolumn{1}{c}{20} 
    & \multicolumn{1}{c}{31.83} 
    & \multicolumn{1}{c}{31.16} 
    & \multicolumn{1}{c}{34.59}
    & \multicolumn{1}{c}{99.98}
    & \multicolumn{1}{c}{18.50}
    & \multicolumn{1}{c}{17.12}  \\
    &					
    & \multicolumn{1}{c}{50} 
    & \multicolumn{1}{c}{31.42} 
    & \multicolumn{1}{c}{30.70} 
    & \multicolumn{1}{c}{34.19}
    & \multicolumn{1}{c}{99.98}
    & \multicolumn{1}{c}{18.08}
    & \multicolumn{1}{c}{17.12}  \\
    &
    & \multicolumn{1}{c}{100} 
    & \multicolumn{1}{c}{30.24} 
    & \multicolumn{1}{c}{29.50} 
    & \multicolumn{1}{c}{33.19}
    & \multicolumn{1}{c}{99.98}
    & \multicolumn{1}{c}{17.98}
    & \multicolumn{1}{c}{16.66}  \\

    \midrule

    \multirow{14}{*}{LLaVA-Med}
    & \multirow{4}{*}{FT-LLM}
    					
    & \multicolumn{1}{c}{10} 
    & \multicolumn{1}{c}{55.22} 
    & \multicolumn{1}{c}{53.64} 
    & \multicolumn{1}{c}{47.01}
    & \multicolumn{1}{c}{3.59}
    & \multicolumn{1}{c}{1.63}
    & \multicolumn{1}{c}{15.27} 
    \\
    &					
    & \multicolumn{1}{c}{20} 
    & \multicolumn{1}{c}{49.36} 
    & \multicolumn{1}{c}{48.09} 
    & \multicolumn{1}{c}{40.71}
    & \multicolumn{1}{c}{3.35}
    & \multicolumn{1}{c}{1.46}
    & \multicolumn{1}{c}{14.01}  \\
    &
    & \multicolumn{1}{c}{50} 
    & \multicolumn{1}{c}{42.34} 
    & \multicolumn{1}{c}{41.07} 
    & \multicolumn{1}{c}{32.51}
    & \multicolumn{1}{c}{2.34}
    & \multicolumn{1}{c}{1.06}
    & \multicolumn{1}{c}{12.28}  \\
    &
    & \multicolumn{1}{c}{100} 
    & \multicolumn{1}{c}{45.58} 
    & \multicolumn{1}{c}{44.21} 
    & \multicolumn{1}{c}{38.11}
    & \multicolumn{1}{c}{0.84}
    & \multicolumn{1}{c}{0.63}
    & \multicolumn{1}{c}{14.55}  \\
    \cmidrule{2-9}

    & \multirow{4}{*}{FT-Proj}
    					
    & \multicolumn{1}{c}{10} 
    & \multicolumn{1}{c}{16.8} 
    & \multicolumn{1}{c}{16.05} 
    & \multicolumn{1}{c}{12.00}
    & \multicolumn{1}{c}{100.0}
    & \multicolumn{1}{c}{0.29}
    & \multicolumn{1}{c}{5.56} 
    \\
    &
    & \multicolumn{1}{c}{20} 
    & \multicolumn{1}{c}{8.96} 
    & \multicolumn{1}{c}{8.31} 
    & \multicolumn{1}{c}{6.29}
    & \multicolumn{1}{c}{100.0}
    & \multicolumn{1}{c}{0.27}
    & \multicolumn{1}{c}{6.10}  \\
    &					
    & \multicolumn{1}{c}{50} 
    & \multicolumn{1}{c}{6.94} 
    & \multicolumn{1}{c}{6.54} 
    & \multicolumn{1}{c}{6.12}
    & \multicolumn{1}{c}{100.0}
    & \multicolumn{1}{c}{0.26}
    & \multicolumn{1}{c}{4.26}  \\
    &
    & \multicolumn{1}{c}{100} 
    & \multicolumn{1}{c}{8.32} 
    & \multicolumn{1}{c}{8.47} 
    & \multicolumn{1}{c}{7.16}
    & \multicolumn{1}{c}{100.0}
    & \multicolumn{1}{c}{0.26}
    & \multicolumn{1}{c}{4.31}  \\

   \cmidrule{2-9}

    & \multirow{4}{*}{SERAC}
    					
    & \multicolumn{1}{c}{10} 
    & \multicolumn{1}{c}{54.23} 
    & \multicolumn{1}{c}{48.51} 
    & \multicolumn{1}{c}{57.51}
    & \multicolumn{1}{c}{99.81}
    & \multicolumn{1}{c}{3.72}
    & \multicolumn{1}{c}{9.77} 
    \\
    &					
    & \multicolumn{1}{c}{20} 
    & \multicolumn{1}{c}{54.31} 
    & \multicolumn{1}{c}{48.55} 
    & \multicolumn{1}{c}{57.58}
    & \multicolumn{1}{c}{99.81}
    & \multicolumn{1}{c}{3.72}
    & \multicolumn{1}{c}{10.15}  \\
    &					
    & \multicolumn{1}{c}{50} 
    & \multicolumn{1}{c}{54.37} 
    & \multicolumn{1}{c}{48.57} 
    & \multicolumn{1}{c}{57.50}
    & \multicolumn{1}{c}{99.81}
    & \multicolumn{1}{c}{3.71}
    & \multicolumn{1}{c}{9.52}  \\
    &				
    & \multicolumn{1}{c}{100} 
    & \multicolumn{1}{c}{54.37} 
    & \multicolumn{1}{c}{48.18} 
    & \multicolumn{1}{c}{57.72}
    & \multicolumn{1}{c}{99.81}
    & \multicolumn{1}{c}{3.68}
    & \multicolumn{1}{c}{9.23}  \\
    \bottomrule
    \end{tabular}
    \label{tab:sequential-edit-results}
\end{table*}

\section{E. More Related Work}

In addition to the knowledge editing methods tested in the paper, there are many other advanced knowledge editing methods, but we did not evaluate them due to the complexity of the methods. Here, we introduce more knowledge editing algorithms.

\paragraph{Additional Parameters}  
This category introduces a small number of trainable parameters to adapt the model to new knowledge without full model finetuning. \textbf{GRACE}~\cite{hartvigsen2023aging} is a lifelong model editing framework designed for streaming error correction of deployed language models. Unlike existing editing methods, which may cause large-scale performance degradation by modifying model weights, GRACE introduces a discrete key-value adapter mechanism: it writes new mappings to the latent space of the pre-trained model, creating a local, editable repair codebook without modifying the original model parameters. This design ensures that edits are only for the target errors, minimizing the impact on irrelevant inputs. \textbf{MELO}~\cite{yu2024melo} is a plug-in model editing method based on dynamic LoRA with neuron index. The core design of this method is to build a neuron index through an internal vector database, dynamically activate specific LoRA modules according to input, and adjust language model behavior without changing the original model structure. This plug-in architecture makes it easy to integrate into a variety of LLM backbone models, while significantly reducing computational overhead. \textbf{WISE}~\cite{wang2024wise} optimizes the knowledge management model in the lifelong editing scenario by reconstructing the storage form, retrieval logic or update rules of knowledge in the model. It is speculated that it may have adopted a more efficient memory organization method (such as dynamic partition storage, associated knowledge index, etc.), so that the model can accurately locate and call the target editing knowledge while avoiding redundant expansion of memory modules. At the same time, by constraining the scope of influence of knowledge updates, the stability of the original knowledge after continuous editing is ensured.

\paragraph{Meta-learning Based}  
This type of algorithm can be considered as a task adaptation problem, which can achieve more efficient and robust editing results by learning "how to learn" or "how to edit". \textbf{InstructEdit}~\cite{zhang2024instructedit} is an instruction-based editing technology. This technology uses simple instructions to enable the editor to adapt to multiple tasks at the same time and build a unified editor for each large language model. It adopts the MEND editing architecture and uses the hypernetwork to implement editing, so that the editor has the ability to understand and apply editing instructions, reduce conflicts between tasks, and ensure that the performance of the multi-task editor on a single task reaches or even exceeds that of the single-task editor. \textbf{MALMEN}~\cite{tanmassive} transforms the parameter offset aggregation problem required for editing into a least squares problem, and updates the model parameters by solving normal equations to ensure accuracy and stability during large-scale editing. At the same time, the algorithm separates the calculation process of the hypernetwork and the basic language model, allowing both to support operations of arbitrary batch sizes under limited memory, breaking through the scale limit of simultaneous editing.

\paragraph{Locate-then-Edit}  
These approaches first locate the target knowledge in model parameters or activations and then apply localized updates. \textbf{Knowledge Neurons (KN)}~\cite{dai2022knowledge} uses BERT as the research object, designs a knowledge attribution method for cloze tasks, and identifies neurons that are strongly related to specific facts. Experiments show that the activation level of such neurons is positively correlated with the accuracy of the expression of the corresponding facts, proving that they are the key carriers of factual knowledge. Based on the positioning results of knowledge neurons, attempts are made to directly edit the target knowledge by adjusting the state of specific knowledge neurons (such as updating or erasing) without full fine-tuning. Although this exploration is still in its early stages, it provides new ideas for "precisely modifying model knowledge and multiple facts with strong locality and generalization guarantees". \textbf{MEMIT}~\cite{mengmass} is a method that supports large-scale memory editing of Transformer models. This method breaks through the previous limitations of editing a single or a small amount of knowledge. The core goal is to achieve "batch knowledge update": by optimizing the model parameter adjustment strategy, the pre-trained language model can simultaneously receive and store a large number of new memories without the need to fully retrain the model.

\section{F. Prompt Template}

We use the API of Deepseek-v3~\cite{liu2024deepseek} to construct all generality, portability, and robustness questions. For multi-hop questions, we first search the knowledge graph based on the edited entities, and provide the retrieved relations and tail entities to the large model in the form of prompt words to generate the final multi-hop question-answering results.

\subsection{F.1 Prompt for Generating Rephrase Questions}

\begin{tcolorbox}
[title=Prompt for Generating Rephrase Questions, label=prompt_template3]
\small

\textbf{[Task]}

Please convert the given sentence into a sentence with the same content but different expressions. Give five answers and return them in the form of a list. Give the answer directly and do not reply with extra content. The reply format is as follows: ['sentence 1', 'sentence 2', 'sentence 3', 'sentence 4', 'sentence 5'] 

\textbf{[Given sentence]}

\end{tcolorbox}

\subsection{F.2 Prompt for Generating Portability Questions}

\begin{tcolorbox}[title=Prompt for Generating 1-hop Portability Questions, label=prompt_template4]
\small

\textbf{[Task]}

Below are the head entity, relation, and tail entity retrieved from the knowledge graph. Please add a question and answer based on this triple. Do not include the name of the head entity in the question and generate a question about the image. For example, "How to treat the disease in the image?" or "Where is the organ in the image located in the body?" The corresponding answer is the provided tail entity. Please provide the generated question and answer directly in \{"question":"", "answer":""\}.

\textbf{[Input triple]} 

\{\"head": "", "relation": "", "tail": ""\}

\end{tcolorbox}


\begin{tcolorbox}[title=Prompt for multi-hop Portability Questions, label=prompt_template1]
\small

\textbf{[Task]} 

Construct multi-hop relations and corresponding visual question answering based on existing medical triples.

\textbf{[Input example]}

Existing 1-hop knowledge (subjective $\rightarrow$ relation $\rightarrow$ objective): diabetic retinopathy $\rightarrow$ Complication $\rightarrow$ blindness
Existing 2-hop knowledge
Corresponding questions and answers: (subjective $\rightarrow$ relation $\rightarrow$ objective): blindness $\rightarrow$ Treatment $\rightarrow$ vitrectomy

\textbf{[Generation requirements]}

1. Construct question-answer pairs based on the provided 1-hop triples and 2-hop triples

2. The corresponding questions for 2-hop should be visual questions, and entity names cannot appear in the questions, and the answers and tail entities must be exactly the same

3. Output the corresponding question-answer pairs for 2-hop in the form of a JSON list

\textbf{[Output format]}

[\{"2-hop question": "\{\}", "2-hop answer": "\{\}" \}]

\textbf{[Input Data]}

\{ "1-hop-subject": "\{\}", "1-hop-relation": "\{\}", "1-hop-object": "\{\}", "2-hop-subject": "\{\}", "2-hop-relation": "\{\}", "2-hop-object": "\{\}",\}

\end{tcolorbox}

\subsection{F.3 Prompt for Generating Adversarial Questions}

\begin{tcolorbox}[title=Prompt for Generating Adversarial Questions, label=prompt_template2]
\small

\textbf{[Task]} 

Generate semantically preserved adversarial examples based on medical image question-answer pairs.

\textbf{[Input example]}

Question: Observe the mammogram. A star-shaped lesion with thin radiating lines. What is this finding?
Answer: spiculated masses

\textbf{[Generation requirements]}

1. Keep the core medical features of the original question (such as "star-shaped", "radiating lines")

2. Add hints and interference by one of the following methods:

- Misleading context: such as patient misrecollection

- Irrelevant clinical details: such as calcification description

- Fuzzy qualifiers: such as "possibly", "somewhat"

- Pseudo-authoritative interference: such as outdated terms

- Symptom confusion: such as incorrect symptom description

3. Ensure that the correct answer remains unchanged, but increase the difficulty of the model output

4. Output in the form of a JSON list, including the original question, adversarial question, attack type, and original answer. The list contains 5 different adversarial samples, and the attack type is not repeated

\textbf{[Output format]}

[
\{"original question": "\{\}", "adversarial question": "\{\}", "attack type": "\{\}", "gold answer": "\{\}"\}
]

\textbf{Input Data}
\{"original question": "\{\}", "gold answer": "\{\}"\}

\end{tcolorbox}

\newpage

\end{document}